\documentclass{article}

\PassOptionsToPackage{numbers, compress}{natbib}

 \usepackage[main, final]{neurips_2026}

\usepackage[utf8]{inputenc} 
\usepackage[T1]{fontenc}    
\usepackage{hyperref}       
\usepackage{url}            
\usepackage{booktabs}       
\usepackage{amsfonts}       
\usepackage{nicefrac}       
\usepackage{microtype}      
\usepackage{xcolor}         

\usepackage{graphicx}
\usepackage{amsmath}
\usepackage[table]{xcolor}
\usepackage{booktabs}
\usepackage{multirow}
\usepackage{enumitem}
\usepackage{mathtools}
\usepackage{tikz}
\usetikzlibrary{arrows.meta}

\definecolor{vcolor}{RGB}{40, 100, 200}    
\definecolor{wcolor}{RGB}{210, 90, 50}     
\definecolor{gcolor}{RGB}{120, 50, 175}    

\usepackage{amsthm}
\usepackage{needspace}
\usepackage{cleveref}
\usepackage{thmtools}
\declaretheorem[name=Theorem]{theorem}
\declaretheorem[name=Lemma]{lemma}
\declaretheorem[name=Assumption]{assumption}
\crefname{assumption}{Assumption}{Assumptions}
\Crefname{assumption}{Assumption}{Assumptions}
\usepackage{makecell}
\usepackage{pifont} 
\usepackage{wrapfig}
\usepackage{svg}
\usepackage{caption}

\crefname{appendix}{Appendix}{Appendices}
\Crefname{appendix}{Appendix}{Appendices}
\crefformat{appendix}{Appendix~#2#1#3}
\AddToHook{cmd/appendix/before}{%
\crefalias{section}{appendix}%
\crefalias{subsection}{appendix}%
\crefalias{subsubsection}{appendix}
}

\newcommand{\best}{\cellcolor[HTML]{A9DFBF}}      
\newcommand{\second}{\cellcolor[HTML]{D4EFDF}}    
\newcommand{\bestavg}{\cellcolor[HTML]{D2B4DE}}   
\newcommand{\secondavg}{\cellcolor[HTML]{E8DAEF}} 

\newcommand{\red}[1]{{\color{red}#1}}

\newcommand{\blue}[1]{{\color{blue}#1}}

\newcommand{\ours}{\text{MVRD}}

\title{Multi-View Relational Distillation for\\Spatial Reasoning with Vision-Language Models}

\author{
    \textbf{Kiet T.~Nguyen\quad Hanbo Shim\quad
    Jinwoo Kim\textsuperscript{\dag}\quad
    Seunghoon Hong\textsuperscript{\dag}} \\
    KAIST \\
}

\begin{document}

\maketitle
\renewcommand{\thefootnote}{\dag}
\footnotetext{Equal advising}
\renewcommand{\thefootnote}{\arabic{footnote}}
\setcounter{footnote}{0}

\vspace{-1em}
\begin{abstract}
Vision-language models (VLMs) have achieved strong image and video understanding, yet their visual-spatial representations remain geometrically fragile, leading to failures in spatial reasoning needed for embodied AI, robotics, and autonomous driving. Prior approaches to geometry grounding either fine-tune VLMs on spatial question answering, which can perpetuate spurious visual representations, or fuse features from large geometry-grounded vision models, which substantially increases model size at inference. Knowledge distillation from geometry-grounded vision models offers an alternative, but directly matching multi-view teacher features can disrupt the pretrained alignment between visual and textual representations, degrading object- and language-semantic capabilities. We propose \textbf{multi-view relational distillation (\ours{})}, which distills patch-wise cosine similarities across views instead of the teacher features themselves. These relations encode geometric correspondences adequate for spatial understanding, while leaving the student representation underdetermined, allowing it to remain close to its pretrained vision-language space. Across representative VLMs, \ours{} improves visual-spatial reasoning, outperforming supervised fine-tuning and feature distillation while approaching feature fusion methods with considerably less added parameters and lower latency. We show that \ours{} makes visual representations more geometric while retaining language alignment, and generalizes to 3D scene understanding tasks such as object grounding, dense captioning, and question answering. Code is available at \url{https://github.com/kietngt00/MVRD}
\end{abstract}

\begin{figure}[h]
    \centering
    \includegraphics[width=0.9\linewidth]{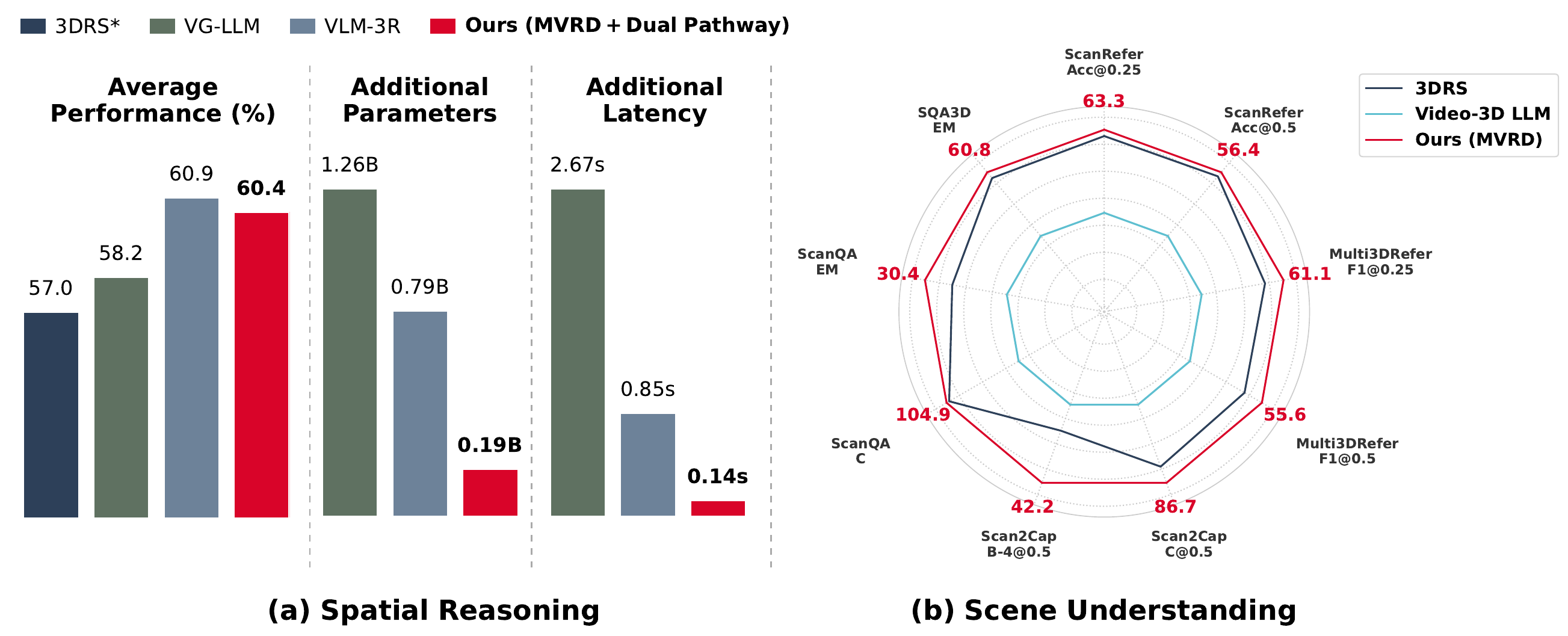}
    \caption{\textbf{Multi-view relational distillation (\ours{}).} 
    \textbf{(a) Spatial reasoning.} \ours{} with Dual Pathway variant approaches the feature fusion method VLM-3R~\cite{fan2025vlm}, using the same backbone and dataset, with far fewer added parameters and lower latency. 
    \textbf{(b) Scene understanding.} \ours{} transfers to scene understanding, surpassing the prior state-of-the-art 3DRS~\cite{huang3DRSMLLMsNeed2025} in all tasks.}
    \label{fig:intro_fig}
\end{figure}

\section{Introduction}
\label{sec:intro}

Humans perceive, understand, and act upon physical environments through rich visual-spatial representations, paired with language as a flexible abstraction for describing scenes, specifying goals, and composing intentions. Evidence from cognitive science and neuroscience suggests that perception and action rely on structured, manipulable representations of visual scenes~\cite{craik1967nature, marr1982vision, johnson1983mental}, supporting capabilities such as object localization~\cite{goodale1992separate}, mental viewpoint changes~\cite{shepard1971mental}, and path planning~\cite{o1978hippocampus}. This complementarity motivates vision-language models (VLMs) that not only align images with text, but also learn geometrically faithful visual representations for grounded spatial reasoning and action, with applications ranging from embodied AI and robotics~\cite{zitkovich2023rt, brohan2022rt, driess2023palm, o2024open} to autonomous driving~\cite{xie2025vlms}.

Although current VLMs are trained with profound scales of data and compute~\cite{bai2025qwen25vltechnicalreport, zhu2025internvl3, zhang2025llavavideo}, it is known that their visual-spatial representations are fractured~\cite{el2024probing}, leading to failures in geometric tasks as simple as inferring relative distances~\cite{li2025sti, yang2025thinking, zhou2025vlm4d}. As such, there has been substantial effort in grounding VLMs to geometry. Early efforts perform supervised fine-tuning (SFT) with spatial question answering~\cite{yang2025visual, yang2025cambrian, cai2025scaling}, which do not directly address the problem of spurious representations~\cite{kumar2025questioning}. More recent work propose to fuse multi-view features from a large geometry-grounded vision model~\cite{hu2025g, zheng2025learning, fan2025vlm, zhang2026spatialstack}, however at the cost of substantially increasing the already large model size during inference time.

Here, we aim to improve spatial representation and understanding in VLMs while minimally increasing their size. Prior work has proposed knowledge distillation~\cite{hinton2015distilling} as a solution, matching multi-view features of VLMs to be similar with those of a geometry-grounded vision model~\cite{huang3DRSMLLMsNeed2025}. We identify a problem of this idea: while it grounds VLMs' visual representations, it obstructs their pretrained alignment with language. We find that this degrades capabilities of linguistic grounding of objects, e.g. tracking the order in which objects of a given class appear. 
Prior work across multiple domains ~\citep{zhang2026videorepa,bhowmik2026moalign,bolya2025perceptionencoderbestvisual,addepalli2024leveraging} has also shown that distillation from off-the-shelf teachers can disrupt a student model’s pretrained representations.
This motivates our key question: \emph{How can we distill geometric knowledge into VLMs without trading off vision-language alignment?}

To this end, we propose \textbf{multi-view relational distillation}, a simple method for improving spatial reasoning in VLMs via knowledge distillation from geometry-grounded vision models. Instead of distilling multi-view features themselves, we distill \emph{cosine similarities} of features across views. Our hypothesis is that multi-view relations encode geometric correspondences adequate to improve spatial understanding of VLMs, yet allow visual representations to remain close to the pretrained vision-language space.
The former is supported by evidence of relational processing in human vision~\cite{li2008unsupervised, wiskott2002slow, whittington2020tolman, tommasi2012psychology}. The latter follows as multi-view similarities do not uniquely determine student features, so there can exist ones with low loss that lie close to the pretraining regime.
Our contributions are as follows:
\begin{enumerate}[label=(\arabic*),leftmargin=*] 
\item \textbf{Failures of feature distillation.} Through evaluations and representation probing, we uncover a previously undiagnosed failure mode of feature distillation, resulting in degradations in vision-language alignment and related tasks despite making VLMs more geometrically grounded.
\item \textbf{Multi-view relational distillation (\ours{}).} Motivated by adequacy and minimality aspects of relational visual representations, we propose \ours{}, an approach that distills similarities within multi-view features from geometry-grounded vision models into VLMs for geometry grounding.
\item \textbf{Competitive performance.} \ours{} outperforms SFT and feature distillation at least by 2.2 points in accuracy on VSI-Bench, and its best setting narrows the gap to the feature fusion counterpart to 0.5 points. We show that \ours{} meaningfully retains vision-language alignment.
\item \textbf{Computational efficiency.} Being a distillation method, \ours{} introduces less than 25\% additional parameters and less than 16\% runtime overhead compared to its feature fusion counterpart.
\end{enumerate}

\section{Related work}

\textbf{VLMs and their fractured spatial understanding.}
VLMs that process text jointly with multiple images, such as multiple views of a scene or video frames, have achieved strong performances in classical benchmarks~\citep{bai2025qwen25vltechnicalreport,zhu2025internvl3,zhang2025llavavideo,comanici2025gemini25pushingfrontier,openai2024gpt4ocard}. However, recent findings, including from VSI-Bench~\citep{yang2025thinking}, have shown that these models are fragile in multi-view spatial reasoning, which requires inferring geometric structures such as egocentric-allocentric transformations~\citep{li2025stibenchmllmsreadyprecise,zhou2025vlm4dspatiotemporalawarenessvision}. This can be attributed to their image encoders~\citep{radford2021learningtransferablevisualmodels,zhai2023sigmoidlosslanguageimage}, which produce representations with limited 3D-awareness needed for depth estimation or matching multi-view correspondences~\citep{el2024probing}. The current paradigm, which fine-tunes VLMs with visual question answering, has been reported to fall short of resolving these spurious visual representations~\citep{tong2024cambrian1fullyopenvisioncentric,asadi2026mirageillusionvisualunderstanding}.

\textbf{Grounding VLMs to visual geometry.} There have been considerable efforts in grounding VLMs to spatial geometry. A line of work proposed to perform SFT on spatial question-answer pairs~\citep{zhang2026flatlandspaceteachingvisionlanguage,yang2025visualspatialtuning,yang2025cambriansspatialsupersensingvideo}.
However, SFT does not directly address the underlying problem of fractured visual representations, and as such, we show in our setting that it alone leads to limited gains.
Recent work tackles this problem at the representation level, leveraging recent large-scale geometry-grounded vision foundation models such as VGGT~\cite{wang2025vggt} and CUT3R~\cite{wang2025cut3r}. \emph{Feature fusion} methods such as VLM-3R~\cite{fan2025vlm}, VG-LLM~\cite{zheng2025learning}, and SpatialStack~\cite{zhangSpatialStack2026} take their features and inject them to the visual pathway of VLMs, with G$^2$VLM~\cite{hu2025g2vlmgeometrygroundedvision} further takes a dual-expert approach, duplicating the model and performing feature fusion in one of them. While these methods demonstrate the benefit of geometry-grounded representations in VLMs, they require a standalone model for feature extraction at inference time, substantially increasing net model size, posing a challenge in embodied applications.

\textbf{Feature distillation methods.}
Given the above, knowledge distillation~\cite{hinton2015distilling} methods that operate at the representation level offer an alternative route towards geometry-grounded VLMs while minimally increasing their size. Importantly, 3DRS~\cite{huang3DRSMLLMsNeed2025} directly aligns multi-view visual features of VLMs with those from the geometry-grounded vision model VGGT. Relatedly, 3DThinker~\cite{chen2026think3dgeometricimagination} and VaLR~\cite{jeon2026visionalignedlatentreasoningmultimodal} align latent chain-of-thought features using curated data and multi-stage training. While 3DRS has been shown to improve geometry grounding of VLMs, and others have gained improvement via chain-of-thought training, how such geometric alignment affects the pretrained language-aligned visual space remains underexplored.
Through careful evaluations and analysis, we find that feature distillation disrupts the alignment between visual and linguistic representations in VLMs, causing degradations in the tasks of estimating the object size or tracking the appearance order of objects.

\textbf{Relational representations and distillation.} Relational representations have been demonstrated to be useful in various contexts \citep{matthews2001short,leinster2014basic, park2019relational}. In human vision, evidence suggest that relational information across temporal stream of visual inputs is leveraged in spatial understanding and object perception \cite{li2008unsupervised, wiskott2002slow, whittington2020tolman, tommasi2012psychology}. In deep learning, a line of work has studied relational knowledge distillation~\citep{park2019relational}, which has been shown to be effective in the vision domain~\cite{dinov3, bolya2025perceptionencoderbestvisual}. 
While these prior work consider relations between images or patches in an image, we extend these approaches and consider relations within multiple views of a scene. Furthermore, we argue that relational distillation better sustains vision-language alignment in VLMs based on the fact that different underlying features can lead to the same relational representation, a property that has not been explicitly leveraged in the prior work.
\section{Failures of geometric feature distillation for VLMs}
\label{sec:problem_setup}


\subsection{Problem setup}
\textbf{VLMs in multi-view setting.} For multi-view scene and video understanding, a VLM takes as input a collection of images ${\bf x}_1, ..., {\bf x}_T\in\mathbb{R}^{H\times W\times 3}$ with a language query ${\bf q}$. A vision encoder maps each image individually into patch-level features, ${\bf x}_t\mapsto {\bf v}_t\in\mathbb{R}^{H_p \times W_p \times d}$, which we refer to as vision tokens. ${\bf q}$ is encoded using subword embeddings from a language model, ${\bf q}\mapsto {\bf w}\in\mathbb{R}^{N\times d}$ with length $N$, which we refer to language tokens. The tokens are concatenated into ${\bf z} = ({\bf v}_1,...,{\bf v}_T, {\bf w})\in\mathbb{R}^{(TH_pW_p+N)\times d}$ and processed by the language model for downstream tasks. At $l$-th layer, we denote the features of these tokens as ${\bf z}^l = ({\bf v}_1^l,...,{\bf v}_T^l, {\bf w}^l)$, where ${\bf z}^0={\bf z}$. Since VLMs tend to focus on aligned use of vision and language, they are often fragile on geometric and spatial tasks. To overcome this, early works performed SFT with spatial question answering, minimizing the loss $\mathcal{L}_{\mathsf{SFT}}\coloneqq -\log p({\bf a}\mid {\bf x}_{1:T},{\bf q})$ where $p$ denotes the likelihood of gold answer ${\bf a}$ evaluated by the VLM.


\textbf{Geometry-grounded features.}
Recent work has proposed to improve spatial understanding in VLMs with representations from a geometry-grounded vision model. Such a model jointly encodes multi-view images into features, ${\bf x}_{1:T}\mapsto {\bf g}_{1:T}\in\mathbb{R}^{T\times H_g \times W_g \times d_g}$, and is trained with specialized objectives such as point cloud and camera pose estimation. As a result, their multi-view features faithfully encode the underlying 3D geometry. Feature fusion methods directly incorporate these into the forward pathway of VLMs, either via resampling ${\bf g}_t\mapsto \bar{\bf g}_t$ into $H_p\times W_p$ grid followed by additions into vision tokens, or through cross-attention. These methods directly address the problem of spurious representations, and yield substantial gains on spatial tasks when combined with SFT. However, they require a standalone vision model for feature extraction, increasing model size.



\textbf{Feature distillation.}
Feature distillation methods train VLMs to match the (resampled) geometric features $\bar{\bf g}_{1:T}$, then discard the additional vision model at inference time, thereby avoiding the overhead of feature fusion. This is done by minimizing the following loss for each input $({\bf x}_{1:T}, {\bf q})$:
\begin{equation}\label{eq:feature_distillation_loss}
\mathcal{L}_{\mathsf{FD}} \coloneqq \frac{1}{TH_pW_p} \sum_{t=1}^T\sum_{h=1}^{H_p}\sum_{w=1}^{W_p} \ell_{\mathsf{FD}}\!\left(f\!\left({\bf v}_t^l[h, w]\right), \,\!\mathsf{sg}(\bar{\bf g}_t[h, w]\right),
\end{equation}
where $\ell_{\mathsf{FD}}:\mathbb{R}^d\times\mathbb{R}^d\to\mathbb{R}$ is a choice of error function, such as cosine distance, $f:\mathbb{R}^{d}\to\mathbb{R}^{d_g}$ is a position-wise MLP projection, and $\mathsf{sg}(\cdot)$ is the stop-gradient operator. The VLM and the projector are trained together with a composite loss $\mathcal{L}_{\mathsf{SFT}} + \lambda \mathcal{L}_{\mathsf{FD}}$ with hyperparameter $\lambda>0$. The composite loss maximizes the likelihood of correct question answering by the VLM, while attempting to make its representations more geometric, by maximizing similarity, aiming for ${\bf v}_t^l\approx \bar{\bf g}_t$.


\subsection{Failures of feature distillation for geometric grounding}
\label{sec:failures_of_feature_distillation}

\begin{table}[t]
\vspace{-2em}
\caption{
\textbf{Feature distillation vs.\ SFT} on \textit{LLaVA-Video-7B-Qwen2}~\citep{zhang2025llavavideo} using VLM-3R~\citep{fan2025vlm} dataset, evaluated on VSI-Bench. Originally reported results are included in Appendix~\ref{appendix:3drs_original}. * indicates training on the VLM-3R dataset.
}
\label{3DRS_and_SFT}
\centering
\renewcommand{\arraystretch}{1.2}
\resizebox{\textwidth}{!}{%
\begin{tabular}{l | c | c c c c c c c c}
\toprule
& &
\rotatebox{70}{Abs. Dist.} & 
\rotatebox{70}{Room Size} & 
\rotatebox{70}{Rel. Dist.} & 
\rotatebox{70}{Rel. Dir.} & 
\rotatebox{70}{Route Plan} & 
\rotatebox{70}{Obj. Count} & 
\rotatebox{70}{Obj. Size} & 
\rotatebox{70}{Appr. Order} \\
Methods & Avg. & 
\multicolumn{5}{c}{\cellcolor[HTML]{FFF2E5}\textbf{Geometric}} & 
\multicolumn{3}{c}{\cellcolor[HTML]{FFFFED}\textbf{Linguistic grounding}} \\
\hline
\rowcolor[HTML]{F2F7FB}
SFT~\cite{zhang2025llavavideo}& \textbf{57.7} & 43.6 & 63.7 & 64.9 & 68.9 & \textbf{40.7} & \textbf{70.6} & \textbf{70.8} & \textbf{38.5} \\
SFT + feature distillation (3DRS*)~\cite{huang3DRSMLLMsNeed2025} & 57.0 & \textbf{51.3} & \textbf{69.1} & \textbf{65.8} & \textbf{74.5} & 37.6 & 70.0 & 68.0 & 19.6 \\
\hline
\end{tabular}}
\end{table}
\begin{figure}[t]
    \centering
    \includegraphics[trim={1cm 0 0 0}, clip, width=\textwidth]{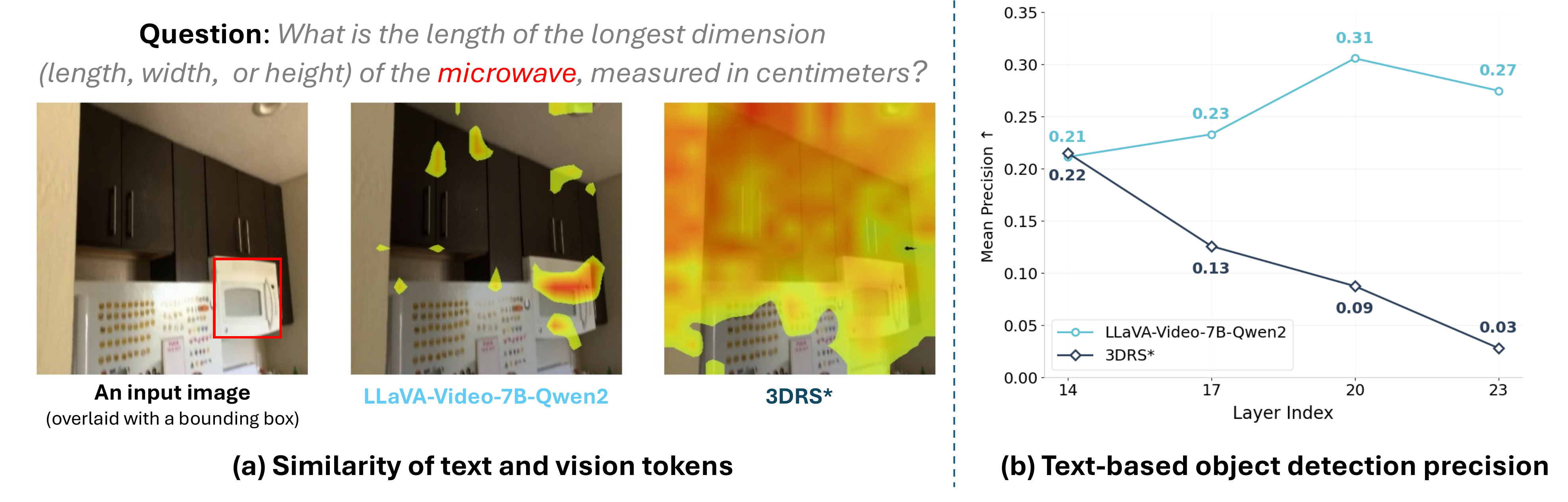}
    \caption[Failure of feature distillation.]{
    \textbf{Failure of feature distillation.}
    \textbf{(a)} \textit{Qualitative}: At layer $20$, 3DRS* fails to align \textit{microwave} with the corresponding visual patches, unlike the base model LLaVA-Video-7B-Qwen2.
    \textbf{(b)} \textit{Quantitative}: A sharp drop in text-based object-detection precision across layers indicates severe degradation in vision-language alignment.
    }
    \vspace{-1em}
    \label{fig:compare_llava_3drs}
\end{figure}

\begin{wrapfigure}[7]{r}{0.2\textwidth}
	\vspace{-2em}
	\centering
	\begin{tikzpicture}[
    vec/.style={-{Stealth[length=5pt,width=4pt]}, line width=0.9pt},
]
 
\draw[vec, vcolor!22] (0,0) -- ({0.62*cos(110)},{0.62*sin(110)});
\draw[vec, vcolor!50] (0,0) -- ({0.85*cos(80)},{0.85*sin(80)});
\draw[vec, vcolor!38] (0,0) -- ({0.74*cos(40)},{0.74*sin(40)});
\draw[vec, vcolor!28] (0,0) -- ({0.66*cos(15)},{0.66*sin(15)});
 
\draw[vec, vcolor!40] (0,0) -- ({0.62*cos(248)},{0.62*sin(248)});
\draw[vec, vcolor!55] (0,0) -- ({0.80*cos(210)},{0.80*sin(210)});
\draw[vec, vcolor!75] (0,0) -- ({0.70*cos(235)},{0.70*sin(235)});
\draw[vec, vcolor]    (0,0) -- ({0.85*cos(220)},{0.85*sin(220)});
\node[vcolor] at ({1.18*cos(225)},{1.18*sin(225)}) {$\mathbf{v}$};
 
\draw[vec, wcolor!50] (0,0) -- ({0.86*cos(258)},{0.86*sin(258)});
\draw[vec, wcolor!70] (0,0) -- ({0.68*cos(282)},{0.68*sin(282)});
\draw[vec, wcolor]    (0,0) -- ({0.80*cos(268)},{0.80*sin(268)});
\node[wcolor] at ({1.12*cos(275)},{1.12*sin(275)}) {$\mathbf{w}$};
 
\draw[vec, gcolor!40] (0,0) -- ({0.65*cos(105)},{0.65*sin(105)});
\draw[vec, gcolor]    (0,0) -- ({0.85*cos(75)},{0.85*sin(75)});
\draw[vec, gcolor!75] (0,0) -- ({0.78*cos(45)},{0.78*sin(45)});
\draw[vec, gcolor!55] (0,0) -- ({0.72*cos(20)},{0.72*sin(20)});
\node[gcolor] at ({1.30*cos(80)},{1.30*sin(80)}) {$\bar{\mathbf{g}}$};
 
\draw[-{Stealth[length=5pt,width=4pt]}, line width=0.8pt, gray!75]
    ({0.6*cos(195)},{0.6*sin(195)})
    arc[start angle=195, end angle=120, radius=0.6];
 
\draw[thin, gray!60] (0,0) circle (1);
\fill (0,0) circle (1.5pt);
 
\end{tikzpicture}
	\label{fig:toy_example}
\end{wrapfigure}
The feature distillation loss in \eqref{eq:feature_distillation_loss} usually enjoys stable optimization properties, and hence achieves the desired grounding ${\bf v}_t^l\approx \bar{\bf g}_t$ with ease. However, we identify its potential problem, obstructing the pretrained alignment between vision and language tokens. We illustrate this problem using a toy example in the figure on the right, showing three sets of feature vectors (\textcolor{vcolor}{${\bf v}$}, \textcolor{wcolor}{${\bf w}$}, \textcolor{gcolor}{$\bar{\bf g}$}) in the two-dimensional plane. As feature distillation (\textcolor{gray!75}{arrow}) brings vision tokens~\textcolor{vcolor}{${\bf v}$} close to geometric features \textcolor{gcolor}{$\bar{\bf g}$}, it drifts away from the original alignment relative to language tokens \textcolor{wcolor}{${\bf w}$}. We posit that this could lead to degradations in pretrained capabilities that jointly use vision \textcolor{vcolor}{${\bf v}$} and language \textcolor{wcolor}{${\bf w}$}, as their associations would become out-of-distribution.

\textbf{Formalization.}
We formalize the above discussion by letting $\mathbf{v},\mathbf{w},\bar{\mathbf{g}}\in\mathbb{R}^{N\times d}$
denote pretrained vision, language, and geometric teacher features
with $\ell_2$-normalized rows~\citep{huang3DRSMLLMsNeed2025}.
Holding $\mathbf{w}$ and $\bar{\mathbf{g}}$ fixed, define \emph{alignment}
$f(\mathbf{v}, \mathbf{w}) :=
\sum_{i=1}^{N} \mathbf{v}_i^\top \mathbf{w}_i$
and initial gaps
$\delta := \|\mathbf{v} - \bar{\mathbf{g}}\|$ and
$\epsilon := \|\mathbf{v} - \mathbf{w}\|$ using the Frobenius norm.
Feature distillation minimizes
$\mathcal{L}_{\mathsf{f}}(\mathbf{x}):=
\ell_{\mathsf{f}}(\mathbf{x},\bar{\mathbf{g}})$
for a distance metric $\ell_{\mathsf{f}}$.

\begin{assumption}
\label{ass:relative_gap}
Motivated by the separate pretraining of the VLM and geometric
teacher, we assume $\delta>a\epsilon$ for some $a>3$.
\end{assumption}

\begin{theorem}
\label{thm:fd_misalignment}
Under \Cref{ass:relative_gap}, feature distillation has minimizer
$\mathbf{v}_{\mathsf{f}}^\ast=\bar{\mathbf{g}}$ and decreases
vision--language alignment:
\begin{equation}
\label{eq:fd_alignment_drop}
    f(\mathbf{v}, \mathbf{w})
    - f(\mathbf{v}_{\mathsf{f}}^\ast, \mathbf{w})
    \geq \tfrac{1}{2}\delta(\delta - 2\epsilon) > 0.
\end{equation}
\end{theorem}

The proof is in \Cref{appendix:proofs}.

\textbf{Demonstration.} We show this failure mode in a large-scale VLM: LLaVA-Video-7B-Qwen2~\citep{zhang2025llavavideo}, by comparing SFT on the VLM3R dataset~\citep{fan2025vlm} with SFT combined with feature distillation following 3DRS \citep{huang3DRSMLLMsNeed2025}, using the representations from VGGT~\citep{wang2025vggt}, a large geometry-grounded vision model. 
For evaluations, we use widely adopted benchmark VSI-Bench~\cite{yang2025thinking}, categorizing its tasks by the primary type of spatial reasoning required: \textbf{(i) geometric-centered} tasks (\textit{Abs. Dist, Room Size, Rel. Dist, Route Plan}), which require aggregating multi-view geometry into an allocentric scene model to reason over inter-object or scene-level relations; \textbf{(ii) linguistic grounding-centered} tasks (\textit{Obj. Count, Obj. Size, Appr. Order}), which require precise grounding of text-queried object categories for object localization or commonsense reasoning. We leave an extended discussion in \Cref{appendix:task_taxonomy}.

In this setting, in \Cref{3DRS_and_SFT} we find that while feature distillation improves geometric tasks, consistent with prior findings \cite{huang3DRSMLLMsNeed2025, chen2026think3dgeometricimagination, jeon2026visionalignedlatentreasoningmultimodal}, it can be \emph{detrimental} to tasks in the second category, which rely on joint use of vision and language such as in per-frame object localization. This implies, like in our toy example, directly grounding features induces a tradeoff in vision-language alignment. The issue was undiagnosed in prior work \cite{huang3DRSMLLMsNeed2025, chen2026think3dgeometricimagination, jeon2026visionalignedlatentreasoningmultimodal} due to their focus on geometry-centric evaluations, such as 3D object grounding or cross-view relation reasoning, which excluded language-semantic tasks.

To understand this problem at the representation level, in \Cref{fig:compare_llava_3drs}(a) we examine the vision-language alignment of the distilled model via cosine similarity scores between vision tokens, and language tokens that refer to a particular object. We observe that, while the vision tokens are grounded to object semantic in SFT, feature distillation indeed loses it, with high scores focused on unrelated but geometrically correlated regions. We further quantify such misalignment by introducing a metric based on language-based object detection precision, comparing the highest-score patches against gold object bounding boxes from ScanNet~\cite{dai2017scannet}. Further details of this metric are given in \Cref{appendix:text-detection}. \Cref{fig:compare_llava_3drs}(b) confirms that the disruption of vision-language alignment happens consistently. 
Similar representation disruption caused by distillation from off-the-shelf teachers has also been observed in prior work across other domains~\citep{zhang2026videorepa,bhowmik2026moalign,bolya2025perceptionencoderbestvisual,addepalli2024leveraging}.
Our finding here motivates the key question: \textit{How can geometric knowledge be distilled into VLMs without compromising vision-language alignment?}

\section{Multi-view relational distillation (\ours{})}
\label{sec:method}

We propose multi-view relational distillation (\ours{}) to address the challenge raised in \Cref{sec:problem_setup}. In \ours{}, instead of distilling features themselves, we distill similarities of features across multiple views of a scene. Our hypothesis is that (i) these relations encode geometric correspondences adequate to improve spatial understanding of VLMs, while (ii) allowing the features themselves to remain close to its pretrained vision-language space.

\begin{wrapfigure}[7]{r}{0.36\textwidth}
	\vspace{-2em}
	\centering
	\begin{tikzpicture}[
    vec/.style={-{Stealth[length=5pt,width=4pt]}, line width=0.9pt},
]

\begin{scope}[shift={(-1.5,0)}]
    \draw[vec, vcolor!40] (0,0) -- ({0.62*cos(248)},{0.62*sin(248)});
    \draw[vec, vcolor!55] (0,0) -- ({0.80*cos(210)},{0.80*sin(210)});
    \draw[vec, vcolor!75] (0,0) -- ({0.70*cos(235)},{0.70*sin(235)});
    \draw[vec, vcolor]    (0,0) -- ({0.85*cos(220)},{0.85*sin(220)});
    \node[vcolor] at ({1.18*cos(225)},{1.18*sin(225)}) {$\mathbf{v}$};

    \draw[vec, wcolor!50] (0,0) -- ({0.86*cos(258)},{0.86*sin(258)});
    \draw[vec, wcolor!70] (0,0) -- ({0.68*cos(282)},{0.68*sin(282)});
    \draw[vec, wcolor]    (0,0) -- ({0.80*cos(268)},{0.80*sin(268)});
    \node[wcolor] at ({1.12*cos(275)},{1.12*sin(275)}) {$\mathbf{w}$};

    \draw[vec, gcolor!40] (0,0) -- ({0.65*cos(105)},{0.65*sin(105)});
    \draw[vec, gcolor]    (0,0) -- ({0.85*cos(75)},{0.85*sin(75)});
    \draw[vec, gcolor!75] (0,0) -- ({0.78*cos(45)},{0.78*sin(45)});
    \draw[vec, gcolor!55] (0,0) -- ({0.72*cos(20)},{0.72*sin(20)});
    \node[gcolor] at ({1.30*cos(80)},{1.30*sin(80)}) {$\bar{\mathbf{g}}$};

    \draw[thin, gray!60] (0,0) circle (1);
    \fill (0,0) circle (1.5pt);
\end{scope}

\begin{scope}[shift={(1.5,0)}]
    \draw[vec, vcolor!55] (0,0) -- ({0.80*cos(185)},{0.80*sin(185)});
    \draw[vec, vcolor!75] (0,0) -- ({0.70*cos(210)},{0.70*sin(210)});
    \draw[vec, vcolor]    (0,0) -- ({0.85*cos(240)},{0.85*sin(240)});
    \draw[vec, vcolor!40] (0,0) -- ({0.62*cos(270)},{0.62*sin(270)});
    \node[vcolor] at ({1.18*cos(225)},{1.18*sin(225)}) {$\mathbf{v}$};

    \draw[vec, wcolor!50] (0,0) -- ({0.86*cos(258)},{0.86*sin(258)});
    \draw[vec, wcolor!70] (0,0) -- ({0.68*cos(282)},{0.68*sin(282)});
    \draw[vec, wcolor]    (0,0) -- ({0.80*cos(268)},{0.80*sin(268)});
    \node[wcolor] at ({1.12*cos(275)},{1.12*sin(275)}) {$\mathbf{w}$};

    \draw[vec, gcolor!40] (0,0) -- ({0.65*cos(105)},{0.65*sin(105)});
    \draw[vec, gcolor]    (0,0) -- ({0.85*cos(75)},{0.85*sin(75)});
    \draw[vec, gcolor!75] (0,0) -- ({0.78*cos(45)},{0.78*sin(45)});
    \draw[vec, gcolor!55] (0,0) -- ({0.72*cos(20)},{0.72*sin(20)});
    \node[gcolor] at ({1.30*cos(80)},{1.30*sin(80)}) {$\bar{\mathbf{g}}$};

    \draw[thin, gray!60] (0,0) circle (1);
    \fill (0,0) circle (1.5pt);
\end{scope}

\draw[-{Stealth[length=5pt,width=4pt]}, line width=0.8pt, gray!75]
    (-0.3,0) -- (0.3,0);

\end{tikzpicture}
\end{wrapfigure}
\textbf{Motivation.} The first part of the hypothesis is motivated by studies on human vision~\cite{li2008unsupervised, wiskott2002slow, whittington2020tolman, tommasi2012psychology}, which suggest that spatial reasoning relies on relational information across a stream of visual inputs. The second part is mathematical, and to illustrate it we recall the two-dimensional example from \Cref{sec:failures_of_feature_distillation} with vision, language, and geometric tokens (\textcolor{vcolor}{${\bf v}$}, \textcolor{wcolor}{${\bf w}$}, \textcolor{gcolor}{$\bar{\bf g}$}), respectively. Instead of matching $\textcolor{vcolor}{{\bf v}}\approx\textcolor{gcolor}{\bar{\bf g}}$ directly, we consider matching pairwise cosine similarities of \textcolor{vcolor}{${\bf v}$} to that of \textcolor{gcolor}{$\bar{\bf g}$}. Importantly, this leaves vision features \emph{underdetermined}, that is, if \textcolor{vcolor}{${\bf v}^*$} is optimal after distillation, then \emph{every} $t(\textcolor{vcolor}{{\bf v}^*})$, where $t$ is any shared rotation and independent scaling of vectors, must be also optimal, as they have the same cosine similarities. This means a learner can choose from these equally geometry-grounded visual representations, favorably one close to its pretrained space, where the alignment between vision \textcolor{vcolor}{${\bf v}$} and language \textcolor{wcolor}{${\bf w}$} is better retained.

Formally, in this motivation, matching pairwise cosine similarities of the row-normalized
features in \Cref{sec:failures_of_feature_distillation} gives
$\mathcal{L}_{\mathsf{r}}(\mathbf{x}):=
\ell_{\mathsf{r}}(\mathbf{x}\mathbf{x}^\top,
\bar{\mathbf{g}}\bar{\mathbf{g}}^\top)$,
where $\ell_{\mathsf{r}}$ is a distance metric on Gram matrices.
This loss is invariant to orthogonal transformations (\Cref{lem:rd_orthogonal_invariance}), admitting equally optimal feature orientations.
Among these optima, we assume the learner selects one close to its pretrained initialization, motivated by the simplicity bias of gradient-based optimization~\citep{shah2020pitfalls}.

\begin{assumption}
\label{ass:rd_near_initialization}
The distillation procedure selects a relational optimum
$\mathbf{v}_{\mathsf{r}}^\ast$ close to the pretrained
initialization $\mathbf{v}$, satisfying
\begin{equation}
\label{eq:rd_proximity}
    \mathcal{L}_{\mathsf{r}}(\mathbf{v}_{\mathsf{r}}^\ast)=0,
    \qquad
    \rho := \|\mathbf{v}_{\mathsf{r}}^\ast-\mathbf{v}\|
    < \epsilon.
\end{equation}
Recall from \Cref{sec:failures_of_feature_distillation} that
$\epsilon=\|\mathbf{v}-\mathbf{w}\|$ denotes the pretrained
vision-language gap.
\end{assumption}

\begin{theorem}
\label{thm:rd_alignment}
Under \Cref{ass:relative_gap,ass:rd_near_initialization},
relational distillation retains better vision--language alignment
than feature distillation:
\begin{equation}
\label{eq:rd_alignment_advantage}
    f(\mathbf{v}_{\mathsf{r}}^\ast,\mathbf{w})
    > f(\mathbf{v}_{\mathsf{f}}^\ast,\mathbf{w}).
\end{equation}
\end{theorem}

The proof is in \Cref{appendix:proofs}.


\textbf{Method.} From these insights, we construct our method as follows. Its core component is \emph{relational representation} $R:\mathbb{R}^{N\times d}\to\mathbb{R}^{N\times N}$, defined upon a choice of similarity function $r:\mathbb{R}^d\times\mathbb{R}^d\to\mathbb{R}$ as $R({\bf h})[i, j] \coloneqq r({\bf h}[i], {\bf h}[j])$. For the vision ${\bf v}_{1:T}^l$ and geometric tokens $\bar{\bf g}_{1:T}$, it defines a multi-view relational representation, with $N = T\times H_p\times W_p$, the latter encoding geometric correspondences, and the goal being its distillation. We therefore define the multi-view relational distillation loss as
\begin{equation}
\mathcal{L}_\mathsf{RD}\coloneqq \ell_\mathsf{RD}\!\left(R\!\left(f({\bf v}_{1:T}^l)\right), R\!\left(\mathsf{sg}(\bar{\bf g}_{1:T}))\right)\right),
\label{eq:relational_distillation}
\end{equation}
where $f:\mathbb{R}^{d}\to\mathbb{R}^{d_g}$ is a position-wise MLP projection, and $\ell_\mathsf{RD}$ is a suitable loss function on pairs of $N\times N$ matrices. The VLM and the MLP projector are trained with a composite loss $\mathcal{L}_\mathsf{SFT}+\gamma\mathcal{L}_\mathsf{RD}$ with hyperparameter $\gamma>0$. It trains VLM on visual question answering, while making its representations more geometric, in the multi-view relational sense determined by the choice of $r$.

\textbf{Choice of similarity.} Our motivating example suggests a natural design criterion for the similarity function $r$. If more features share the same pairwise similarities, there is more freedom to select visual features that remain close to the pretrained space. Cosine similarity is suitable in this regard because it is invariant to feature magnitudes and global orientation of the feature set. Thus, relational distillation only need match relative angles, leaving global orientation and per-feature scales largely unconstrained. By contrast, negative Euclidean distance lacks scale invariance, forcing distillation to also match feature scales. \Cref{appendix:relation-ablation} verifies that cosine similarity performs better than Euclidean.

\textbf{Choice of matrix loss.} In prior work on relational distillation, mean squared error (MSE) has been used as the loss $\ell_\mathsf{RD}$ on relational representation matrices. We empirically find it better to use negative Person correlation computed row-wise and averaged. This is partially since the resulting loss $\ell_\mathsf{RD}$ is invariant to the scales of the similarity scores, which improves its robustness across scales, reducing the need to carefully tune the hyperparameters for training stability. In addition to this choice, to adhere to causal attention of VLMs, we apply causal masking before computing the loss, as well as zeroing the diagonal to remove the role of trivial self-similarity. \Cref{appendix:relation-ablation} verifies that the correlation loss performs better than MSE.

\begin{figure}
    \centering
    \includegraphics[width=0.9\linewidth]{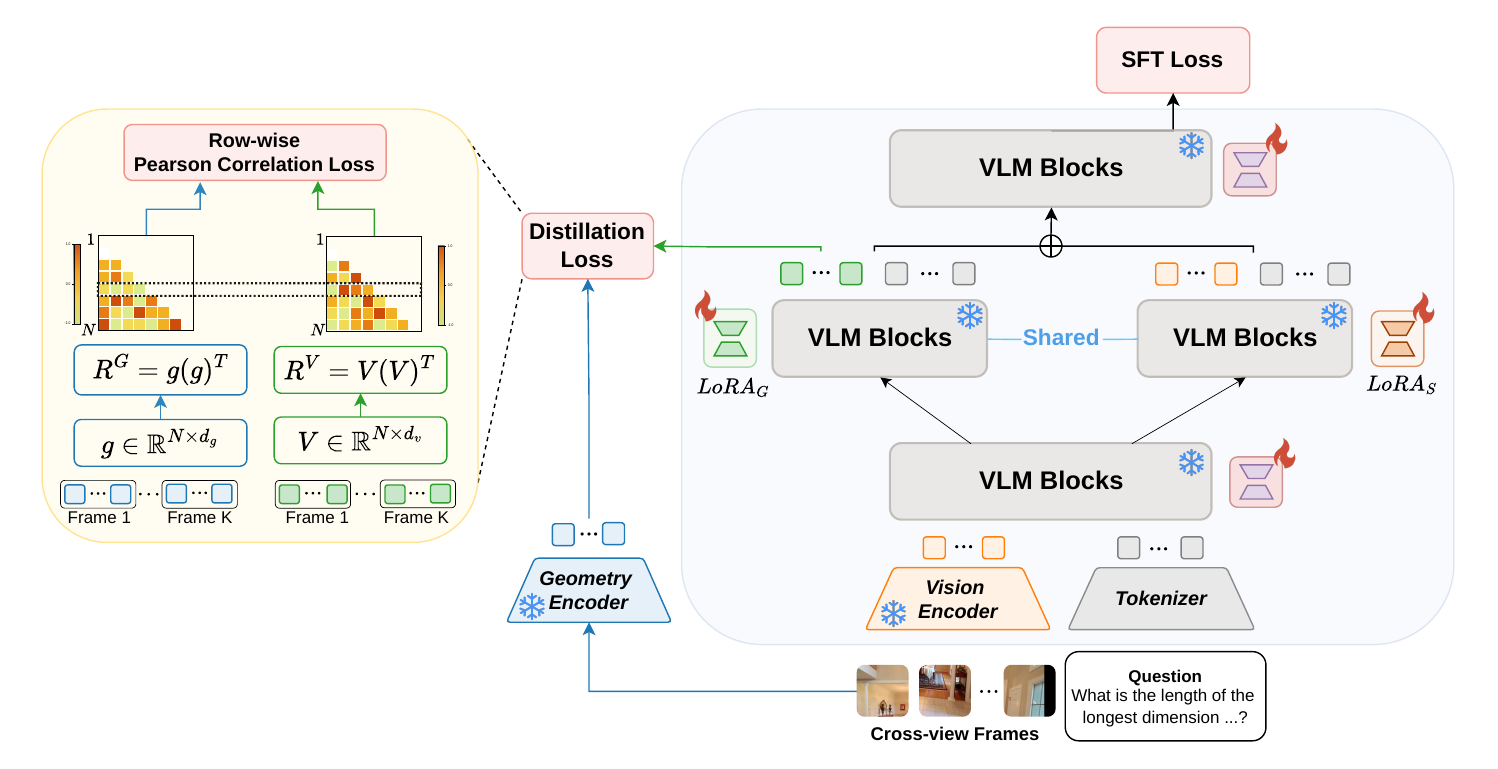}
    \caption{
    An overview of multi-view relational distillation loss and dual pathway.
    }
    \label{fig:architecture}
\vspace{-1em}
\end{figure}



\textbf{Architectural aspects.} To further minimize the drift from the pretrained representation space, we use low-rank adaptation (LoRA)~\cite{hu2021loralowrankadaptationlarge}. Yet, as motivated in the beginning of the section, jointly supervising a shared parameter set with $\mathcal{L}_{\mathsf{SFT}}$ and $\mathcal{L}_{\mathsf{RD}}$ may still perturb the language-relevant subspace. We therefore decouple the two objectives through a lightweight parameter separation we refer to as \textbf{Dual Pathway} (\Cref{fig:architecture}): at several intermediate layers, a \textit{geometric adapter}, a LoRA module supervised by the composite loss, learns from multi-view relational signal, while a parallel \textit{semantic adapter} supervised only by $\mathcal{L}_{\mathsf{SFT}}$ preserves language-aligned behavior. Their outputs are summed after the distillation layer and passed through the remaining VLM blocks. With only modest parameter and latency overhead, this design nearly matches the feature-fusion baseline, as shown in the next section.
\section{Experiments}

\subsection{Main evaluation: Visual-spatial reasoning}
\label{sec:vsi-experiments}
\textbf{Datasets and evaluation.}
We train on the spatial reasoning dataset released by VLM-3R~\cite{fan2025vlm}, which contains over 200{,}000 question--answer pairs grounded in scenes from ScanNet~\cite{dai2017scannet}, ScanNet++~\cite{yeshwanth2023scannethighfidelitydataset3d}, and ARKitScenes~\cite{baruch2022arkitscenesdiverserealworlddataset}. This dataset has been shown to be effective for learning spatial reasoning~\citep{fan2025vlm,cai2026scalingspatialintelligencemultimodal,zheng2025learning,yang2025visualspatialtuning} while remaining tractable enough to support ablations across diverse settings. We evaluate on VSI-Bench~\cite{yang2025thinking}, which comprises roughly $5{,}000$ QA pairs spanning eight spatial reasoning tasks, and follow its official evaluation protocol for both Multiple-Choice and Numerical-Answer formats. For both training and evaluation, we sample $32$ frames per scene video as multi-view input. To mitigate overfitting when distilling from a geometric teacher tied to fixed frame indices, we apply a frame-sampling augmentation detailed in \Cref{appendix:hyperparams-spatial}.

\textbf{Baselines.}
We compare against three groups of models. 
{(\bf 1)} \emph{Proprietary VLMs}, including GPT-4o~\cite{openai2024gpt4ocard} and Gemini-2.5-Pro~\cite{comanici2025gemini25pushingfrontier}. 
{(\bf 2)} \emph{Open-source general-purpose VLMs}, including InternVL3~\cite{zhu2025internvl3}, LLaVA-Video~\cite{zhang2025llavavideo}, LLaVA-OneVision~\cite{li2024llavaonevisioneasyvisualtask}, and Qwen2.5-VL~\cite{bai2025qwen25vltechnicalreport}. 
{(\bf 3)} \emph{Spatial Reasoning VLMs}, including VLM-3R~\cite{fan2025vlm}, VG-LLM~\cite{zheng2025learning}, 3DRS~\cite{huang3DRSMLLMsNeed2025}, and LLaVA-Video~\cite{zhang2025llavavideo}.
For the last category, we only include models fine-tuned on the same training data as ours, which allows us to better isolate methodological contributions from the effect of additional data.
Also, all baselines except VG-LLM are based on the same VLM backbone.
\blue{
}

\textbf{Implementation details.}
Our model is based on LLaVA-Video-7B-Qwen2~\cite{zhang2025llavavideo} (LLaVA-Video-7B hereafter), with VGGT~\cite{wang2025vggt} as the frozen geometric teacher. 
Distillation is performed at $20$th layer of VLM backbone (0-indexed; $28$ layers total); 
our dual-pathway variant splits LoRA~\cite{hu2021loralowrankadaptationlarge} adapters into geometric and semantic pathways over layers $13$--$20$. 
We fine-tune the LoRA adapters (rank $128$, $\alpha=256$), the input projection of the vision tokens in VLM, and the distillation MLP while keeping the rest frozen. 
Training runs on 8 NVIDIA H200 GPUs. Full hyperparameters are in \Cref{appendix:hyperparams-spatial}.

\begin{table*}[t]

\newcommand{\yesGeo}{\textcolor{black}{\ding{51}}}
\newcommand{\noGeo}{\textcolor{black!35}{\ding{55}}}
\caption{\textbf{Evaluation on VSI-Bench.} 
{\setlength{\fboxsep}{1pt}\colorbox[HTML]{A9DFBF}{dark green}} and 
{\setlength{\fboxsep}{1pt}\colorbox[HTML]{D4EFDF}{light green}} indicate the first and second best open-source results per task column, respectively. Within models' average accuracy,
{\setlength{\fboxsep}{1pt}\colorbox[HTML]{D1BBDE}{dark purple}} and 
{\setlength{\fboxsep}{1pt}\colorbox[HTML]{E8DAEF}{light purple}} are used to denote first and second best performance. \textbf{\ours{} + Dual Pathway} approaches feature-fusion VLM-3R without using additonal geometric encoder.}

\label{table:vsibench_baselines_comparison}
\centering
\renewcommand{\arraystretch}{1.2}
\setlength{\tabcolsep}{6pt}
\resizebox{0.9\textwidth}{!}{%
\begin{tabular}{l | c | c | c c c c c c c c}
\toprule
\multirow{2}{*}{Methods} & Geo. & \multirow{2}{*}{Avg.} &
\rotatebox{70}{Abs. Dist.} &
\rotatebox{70}{Room Size} &
\rotatebox{70}{Rel. Dist.} &
\rotatebox{70}{Rel. Dir.}  &
\rotatebox{70}{Route Plan}&
\rotatebox{70}{Obj. Count} &
\rotatebox{70}{Obj. Size}  &
\rotatebox{70}{Appr. Order}\\
& Enc. & &
\multicolumn{5}{c}{\cellcolor[HTML]{FFF2E5}\textbf{Geometric}} &
\multicolumn{3}{c}{\cellcolor[HTML]{FFFFED}\textbf{Linguistic grounding}} \\
\midrule

\rowcolor[HTML]{F2F7FB}
\multicolumn{11}{l}{\textit{\textbf{Baselines}}} \\
Human~\cite{yang2025thinking} & -- & 79.2 & 47.0 & 45.9 & 94.7 & 95.8 & 95.8  & 94.3 & 60.4 & 100.0 \\
Random Choice                 & -- & --   & --   & --   & 25.0 & 47.9 & 28.4  & --   & --   & 25.2 \\
\midrule

\rowcolor[HTML]{F2F7FB}
\multicolumn{11}{l}{\textit{\textbf{Proprietary VLMs}}} \\
GPT-4o~\cite{openai2024gpt4ocard}                         & \noGeo & 34.0 & 5.3  & 38.2 & 37.0 & 41.3 & 31.5 & 46.2 & 43.8 & 28.5 \\
Gemini-2.5 Pro~\cite{comanici2025gemini25pushingfrontier} & \noGeo & 51.5 & 34.9 & 42.8 & 61.1 & 47.8 & 45.9 & 43.8 & 64.3 & 71.3 \\
\midrule

\rowcolor[HTML]{F2F7FB}
\multicolumn{11}{l}{\textit{\textbf{Open-source general-purpose VLMs}}} \\
Qwen2.5-VL-7B-Instruct~\cite{bai2025qwen25vltechnicalreport}    & \noGeo & 29.3 & 10.5 & 29.6 & 38.4 & 38.0 & 29.8 & 25.2 & 36.4 & 26.8 \\
LLaVA-OneVision-7B~\cite{li2024llavaonevisioneasyvisualtask}    & \noGeo & 32.4 & 20.2 & 12.3 & 42.5 & 35.2 & 29.4 & 47.7 & 47.4 & 24.4 \\
LLaVA-Video-7B~\cite{zhang2025llavavideo} & \noGeo & 35.6 & 14.0 & 24.2 & 43.5 & 42.4 & 34.0 & 48.5 & 47.8 & 30.6 \\
LLaVA-OneVision-72B~\cite{li2024llavaonevisioneasyvisualtask}   & \noGeo & 40.2 & 23.9 & 37.5 & 42.5 & 39.9 & 32.5 & 43.5 & 57.6 & 44.6 \\
LLaVA-Video-72B~\cite{zhang2025llavavideo}& \noGeo & 40.9 & 22.8 & 35.3 & 42.4 & 36.7 & 35.0 & 48.9 & 57.4 & 48.6 \\
InternVL3-8B~\cite{zhu2025internvl3}  & \noGeo & 42.1 & 39.0 & 33.6 & 48.3 & 36.4 & 27.3 & 68.1 & 48.4 & 35.4 \\
InternVL3-78B~\cite{zhu2025internvl3} & \noGeo & 48.4 & 53.7 & 39.5 & 55.9 & 39.5 & 28.9 & 71.2 & 44.4 & 54.5 \\
\midrule

\rowcolor[HTML]{F2F7FB}
\multicolumn{11}{l}{\textit{\textbf{Spatial Reasoning VLMs}}} \\
VG-LLM-8B~\cite{zheng2025learning} & \yesGeo & 58.2          & 53.8 & 62.1 & 63.8 & 83.0 & 44.3 & 71.7 & 68.8 & 18.4 \\
VLM-3R~\cite{fan2025vlm}           & \yesGeo & \bestavg 60.9 & 49.4 & 67.1 & 65.4 & 80.5 & 45.4 & 70.2 & 69.2 & 40.1 \\
\cmidrule(lr){2-11}
LLaVA-Video-7B (\textit{Fine-tuned})~\cite{zhang2025llavavideo} & \noGeo & 57.7            & 43.6         & 63.7         & \second 64.9 & 68.9         & \second 40.7 & 70.6         & \best 70.8   & \best 38.5   \\
3DRS*~\cite{huang3DRSMLLMsNeed2025}    & \noGeo & 57.0              & \best 51.3   & \second 69.1 & \best 65.8   & 74.5         & 37.6         & 70.0         & 68.0         & 19.6         \\
\textbf{\ours{}}                      & \noGeo & 59.9              & \second 50.5 & \best 69.3   & 62.7         & \best 80.0   & \second 40.7 & \best 72.9   & 68.3         & 35.1         \\
\textbf{\ours{} + Dual Pathway}                 & \noGeo & \secondavg 60.4   & 49.6         & 68.6         & 64.4         & \second 78.2 & \best 42.8   & \second 71.8 & \second 70.1 & \second 37.7 \\
\bottomrule
\end{tabular}%
}
\vspace{-0.5em}
\end{table*}
\textbf{Comparison on VSI-Bench.}
We evaluate the spatial reasoning capability of our model on VSI-Bench~\cite{yang2025thinking} in \Cref{table:vsibench_baselines_comparison}.
Comparison with feature distillation - 3DRS~\cite{huang3DRSMLLMsNeed2025}, \ours{} improves on 6 of 8 VSI-Bench tasks. The gains are most consistent on the linguistic-grounding task group, where \ours{} wins on every task and 3DRS are considerably below the supervised fine-tuned (SFT) LLaVA-Video baseline. This indicates that our relational distillation target reduces vision--language misalignment, suffers from a smaller performance drop on tasks that rely on linguistic object localization, where prior feature distillation degrades the underlying VLM.

At the same time, our relational target injects enough geometric structure to clearly surpass the SFT baseline on three geometric-centered tasks: Abs.~Dist., Room Size, and most prominently Rel.~Dir., where \ours{} exceeds SFT by \textbf{+11.1\%} and 3DRS* by \textbf{+5.5\%}. Aggregated over all tasks, \ours{} reaches an average accuracy of $59.9\%$, surpassing the feature-fusion method VG-LLM-8B~\cite{zheng2025learning}, when trained on the same data.

Moreover, decoupling the geometric and semantic pathways (\ours{} + Dual Pathways) further preserves vision--language alignment: accuracy on Appr.~Order and Obj.~Size closely tracks (and on Obj.~Size slightly exceeds) the SFT model. This trades a small drop on geometric-centered tasks for a more balanced overall performance, yielding an average of \textbf{60.4\%}, within $0.5$ points of the state-of-the-art feature-fusion method VLM-3R~\cite{fan2025vlm} trained on the same data. As shown in Figure~\ref{fig:intro_fig}, this comes with only modest additional parameters and runtime overhead relative to VLM-3R.

\textbf{Computation analysis.}
We examine the inference-time overhead of \ours{} + Dual Pathway relative to
feature-fusion methods that rely on an auxiliary geometric encoder at test
time: VLM-3R~\citep{fan2025vlm} (CUT3R~\citep{wang2025cut3r}) and
VG-LLM~\citep{zheng2025learning} (VGGT~\citep{wang2025vggt}). Under matched
inputs of 32 views on a single H200 GPU,
Figure~\ref{fig:intro_fig} shows that dual pathway introduces only modest cost
compared to integrating a full geometric encoder. Specifically, \ours{} + Dual Pathway
adds 0.19B parameters - roughly \textbf{25\%} of CUT3R and
\textbf{15\%} of VGGT - and incurs only \textbf{16\%} of CUT3R's and
\textbf{5\%} of VGGT's added latency. Combined with the accuracy reported
in \Cref{table:vsibench_baselines_comparison}, this shows that
\ours{} + Dual Pathway approaches feature-fusion VLM-3R's performance on VSI-Bench at a fraction of the inference overhead.


\subsection{Analysis}

\textbf{Distillation loss.}
\label{sec:distillation_loss_ablation}
For deeper analysis on improvement, we fix all training settings and compare feature distillation with relation distillation. 
As shown in \Cref{table:ablation_representation_relation}, \ours{} consistently improves the average score over the corresponding feature distillation baseline, both with and without dual pathway. 
It suggests that relation distillation provides a stronger geometric learning signal while better preserving vision-language alignment. 
We also find that LoRA-based feature distillation outperforms the fully fine-tuned 3DRS$^{*}$~\cite{huang3DRSMLLMsNeed2025}, indicating that LoRA helps mitigate catastrophic forgetting.

\begin{table}[t]
\newcommand{\yesPS}{\textcolor{black}{\ding{51}}}
\newcommand{\noPS}{\textcolor{black!35}{\ding{55}}}
\caption{Ablation on distillation target and dual pathway. \emph{Feat Dist} means feature distillation.}
\label{table:ablation_representation_relation}
\centering
\renewcommand{\arraystretch}{1}
\setlength{\tabcolsep}{4pt}
\resizebox{\textwidth}{!}{%
\begin{tabular}{l | c c c | c | c c c c c c c c}
\toprule
& & & & &
\rotatebox{60}{Abs. Dist.} &
\rotatebox{60}{Room Size} &
\rotatebox{60}{Rel. Dist.} &
\rotatebox{60}{Rel. Dir.}  &
\rotatebox{60}{Route Plan}&
\rotatebox{60}{Obj. Count}&
\rotatebox{60}{Obj. Size}  &
\rotatebox{60}{Appr. Order} \\
\textbf{Method} &
\shortstack{\textbf{Distillation}\\\textbf{Target}} &
\shortstack{\textbf{Trainable}\\\textbf{Params}} &
\shortstack{\textbf{Dual}\\\textbf{Path}} &
\textbf{Avg.} &
\multicolumn{5}{c}{\cellcolor[HTML]{FFF2E5}\textbf{Geometric}} &
\multicolumn{3}{c}{\cellcolor[HTML]{FFFFED}\textbf{Linguistic grounding}} \\
\midrule
3DRS$^{*}$       & feature & Full & --     & 57.0            & 51.3         & 69.1         & 65.8         & 74.5         & 37.6         & 70.0         & 68.0         & 19.6 \\
\midrule
Feat Dist   & feature & LoRA & \noPS  & 58.5            & 48.3         & 68.4         & 61.4         & 75.7         & \best 44.9   & 71.4         & 67.8         & 30.1 \\
\textbf{\ours{}}    & relation  & LoRA & \noPS  & \best 59.9 & \best 50.5   & \best 69.3   & \best 62.7         & \best 80.0   & 40.7         & \best 72.9   & \best 68.3         & \best 35.1 \\
\cmidrule(lr){5-13}  
Feat Dist + Dual Pathway   & feature & LoRA & \yesPS & 59.3            & 49.0         & 61.7         & \best 67.3   & 76.4         & \best 43.8   & 71.3         & 69.5         & 35.8 \\
\textbf{\ours{} + Dual Pathway}    & relation  & LoRA & \yesPS & \best 60.4   & \best 49.6   & \best 68.6   & 64.4         & \best 78.2   & 42.8         & \best 71.8   & \best 70.1   & \best 37.7 \\
\bottomrule
\end{tabular}%
}
\vspace{-1em}
\end{table} 

\textbf{Evaluation on general VQA benchmarks.}
To evaluate transfer of \ours{} to general visual question answering (VQA), we extend the controlled comparison (\Cref{table:ablation_representation_relation}) to the multi-image subsets
of BLINK~\cite{fu2024blink} and MuirBench~\cite{wang2025muirbench}
(denoted BLINK* and MuirBench*). These benchmarks assess understanding
across multiple images through tasks such as art-style matching,
forensic detection, allowing us to
examine transfer beyond multi-view scene geometry.
As shown in \Cref{table:additional_vqa}, which also includes VSI-Bench,
\ours{} improves all three scores over feature distillation without
Dual Pathway, raising the average from $45.5$ to $47.3$.
Dual Pathway further improves all three scores for both distillation
targets. With Dual Pathway, \ours{} outperforms feature distillation
on VSI-Bench and BLINK*, while scoring $0.1$ points lower on MuirBench*,
yielding a higher overall average.
These results demonstrate benefits of both relational distillation
and Dual Pathway beyond VSI-Bench.

\begin{table}[t]
\caption{Evaluation on VSI-Bench and additional general VQA benchmarks.}
\label{table:additional_vqa}
\centering
\setlength{\tabcolsep}{6pt}
\renewcommand{\arraystretch}{1.1}
\resizebox{0.95\textwidth}{!}{%
\begin{tabular}{l | c c c | c | c c c}
\toprule
\textbf{Method} &
\shortstack{\textbf{Distillation}\\\textbf{Target}} &
\shortstack{\textbf{Trainable}\\\textbf{Params}} &
\shortstack{\textbf{Dual}\\\textbf{Path}} &
\textbf{Avg.} & \textbf{VSI-Bench} & \textbf{BLINK*} &
\textbf{MuirBench*} \\
\midrule
Feat Dist & feature & LoRA & \textcolor{black!35}{\ding{55}} &
45.5 & 58.5 & 36.8 & 41.1 \\
\textbf{\ours{}} & relation & LoRA & \textcolor{black!35}{\ding{55}} &
\best 47.3 & \best 59.9 & \best 40.3 & \best 41.6 \\
\cmidrule(lr){5-8}
Feat Dist + Dual Pathway & feature & LoRA & \textcolor{black}{\ding{51}} &
47.6 & 59.3 & 39.8 & \best 43.7 \\
\textbf{\ours{} + Dual Pathway} & relation & LoRA & \textcolor{black}{\ding{51}} &
\best 48.3 & \best 60.4 & \best 40.9 & 43.6 \\
\bottomrule
\end{tabular}
}
\vspace{-0.5em}
\end{table}

\begin{figure}[t!]
    \centering
    \includegraphics[trim={0 0 0 0}, clip, width=0.95\textwidth]{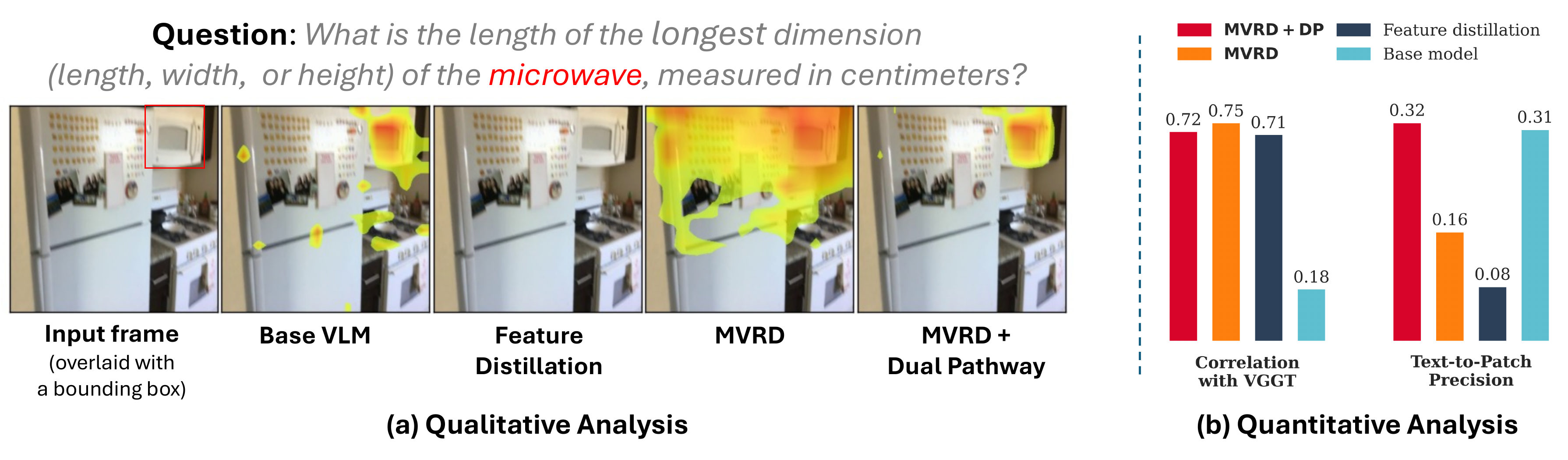}
    \caption{
    \textbf{Vision-language alignment across methods.} 
    \textbf{(a)} At distillation layer, \ours{} and \ours{} + Dual Pathway align \textit{microwave} with its visual patches better than feature distillation. 
    \textbf{(b)} Both improve vision-language alignment while matching feature distillation in geometric relation modeling.
    }
    \label{fig:compare_relation_feature_distill}
\vspace{-1em}
\end{figure}

\textbf{Analysis on visual representation.}
\label{sec:visual_rep_analysis}
To better understand the representation space induced by distillation, we provide a qualitative example in \Cref{fig:compare_relation_feature_distill}(a); additional multi-frame examples are deferred to \Cref{appendix:qualitative}. 
At the distillation layer, feature distillation
visibly degrades text--vision alignment: object text tokens assign low cosine similarity to their corresponding visual patches.
In contrast, \ours{} yields high similarity to both the corresponding patch and its neighbors. We attribute this to relation learning from the geometric teacher: since local patches encode local 3D regions and the teacher's multi-view relations reflect physical-world proximity (\Cref{appendix:vggt-vs-3d}, \Cref{fig:correlation_vggt_3d_distance}), the VLM's patch features inherit this proximity, spreading text--patch similarity to neighboring regions. \ours{} + Dual Pathway resolves this by decoupling multi-view relation learning from the pre-trained vision--language alignment, yielding precise text--patch correspondence.

We further quantify this effect along two complementary axes, both computed at the distillation layer on a subset of ScanNet validation scenes.
\emph{(i)} geometric relation is measured by $-\mathcal{L}_\mathsf{RD}$, i.e. the row-wise Pearson correlation between VLM's relation representation $R({\bf v}_{1:T}^l)$ and VGGT's relation representation $R\!\left(\bar{\bf g}\right)$;
\emph{(ii)} Vision--language alignment is measured by text-grounded object detection precision.
As shown in \Cref{fig:compare_relation_feature_distill}(b), both distillation approaches capture geometric relations, but \ours{} better preserves vision-language alignment than feature distillation, achieving 0.16 vs. 0.08 detection precision. With efficient parameter separation,
\ours{} + Dual Pathway achieves both properties, explaining its leading average accuracy among distillation settings in \Cref{table:vsibench_baselines_comparison}.

\begin{table}[t]
\renewcommand{\arraystretch}{1.25}
\setlength{\tabcolsep}{6pt}
\centering
\begin{minipage}[t]{0.46\textwidth}
  \centering
  \caption{Distillation Layer Ablation.}
  \label{table:distill_layer}
  \resizebox{\linewidth}{!}{%
  \begin{tabular}{l | c c c c | c}
  \toprule
  \textbf{Distill. Layer} & \textbf{L7} & \textbf{L14} & \textbf{L20} & \textbf{Last} & \textbf{SFT-only} \\
  \midrule
  Overall Score & 53.9 & 57.3 & \best 59.9 & 59.8 & 57.7 \\
  \bottomrule
  \end{tabular}%
  }
\end{minipage}
\hfill
\begin{minipage}[t]{0.49\textwidth}
  \centering
  \caption{Separation Layer Selection Ablation.}
  \label{table:layer_separation}
  \resizebox{\linewidth}{!}{%
  \begin{tabular}{l | c c c c c}
  \toprule
  \textbf{Layers for Sep.} & \textbf{0--20} & \textbf{7--20} & \textbf{10--20} & \textbf{13--20} & \textbf{16--20} \\
  \midrule
  Overall Score & 60.2 & 59.8 & 60.0 & \best 60.4 & 59.5 \\
  \bottomrule
  \end{tabular}%
  }
\end{minipage}
\vspace{-1em}
\end{table}

\begin{table}[!t]
\caption{Performance comparison across different VLM backbones.}
\label{table:different_vlms}
\centering
\setlength{\tabcolsep}{5pt}
\resizebox{0.7\textwidth}{!}{%
\begin{tabular}{l | c | c | c}
\toprule
\textbf{Training Method} & \textbf{LLaVA-Video-7B} & \textbf{InternVL3-8B} & \textbf{Qwen2.5-VL-7B} \\
\midrule
\textit{Base (no fine-tuning)}    & 35.6          & 42.1          & 29.3 \\
\midrule
SFT                               & 57.7          & 50.3          & \second 56.1 \\
Feature Distillation       & 58.5          & 57.0          & 55.5 \\
\midrule
\textbf{\ours{}}                  & \second 59.9  & \second 58.7  & \second 56.1 \\
\textbf{\ours{} + Dual Pathway}             & \best 60.4    & \best 59.9    & \best 59.8 \\
\bottomrule
\end{tabular}%
}
\vspace{-1em}
\end{table}

\begin{table}[!t]
\newcommand{\yesGeo}{\textcolor{black}{\ding{51}}}
\newcommand{\noGeo}{\textcolor{black!35}{\ding{55}}}
\caption{Results on 3D scene understanding}
\label{table:main_scene_understanding}
\centering
\resizebox{\textwidth}{!}{%
\begin{tabular}{l c c c c c c c c c c}
\toprule
\multirow{2}{*}{Method} & \multirow{2}{*}{3D Encoder} & \multicolumn{2}{c}{ScanRefer} & \multicolumn{2}{c}{Multi3DRefer} & \multicolumn{2}{c}{Scan2Cap} & \multicolumn{2}{c}{ScanQA} & \multicolumn{1}{c}{SQA3D} \\
\cmidrule(lr){3-4} \cmidrule(lr){5-6} \cmidrule(lr){7-8} \cmidrule(lr){9-10} \cmidrule(lr){11-11}
 & & Acc@0.25 & Acc@0.5 & F1@0.25 & F1@0.5 & C@0.5 & B-4@0.5 & C & EM & EM \\
\midrule
\rowcolor[HTML]{F2F7FB}
\multicolumn{11}{l}{\textit{Geometry-based Methods}} \\
\midrule
Grounded 3D-LLM~\cite{chen2024grounded3dllmreferenttokens} & \yesGeo & 47.9 & 44.1 & 45.2 & 40.6 & 70.6 & 35.5 & 72.7 & - & - \\
PQ3D~\cite{zhu2024unifying3dvisionlanguageunderstanding} & \yesGeo & 57.0 & 51.2 & - & 50.1 & 80.3 & 36.0 & - & - & 47.1 \\
ChatScene~\cite{huang2024chatscenebridging3dscene} & \yesGeo & 55.5 & 50.2 & 57.1 & 52.4 & 77.1 & 36.3 & 87.7 & 21.6 & 54.6 \\
Inst3D-LLM~\cite{yu2025inst3dlmminstanceaware3dscene} & \yesGeo & 57.8 & 51.6 & 58.3 & 53.5 & 79.7 & 38.3 & 88.6 & 24.6 & - \\
3D-LLaVA~\cite{deng20253dllavageneralist3dlmms} & \yesGeo & 51.2 & 40.6 & - & - & 78.8 & 36.9 & 92.6 & - & 54.5 \\
\midrule
\rowcolor[HTML]{F2F7FB}
\multicolumn{11}{l}{\textit{View-based Methods}} \\
\midrule
LLaVA-3D~\cite{zhu2025llava3dsimpleeffectivepathway} & \noGeo & 54.1 & 42.4 & - & - & 79.2 & 41.1 & 91.7 & 27.0 & 55.6 \\
Video-3D LLM~\cite{zheng2025video3dllmlearningpositionaware} & \noGeo & 58.1 & 51.7 & 58.0 & 52.7 & 83.8 & 41.3 & 102.1 & 30.1 & 58.6 \\
Ross3D~\cite{wang2025ross3dreconstructivevisualinstruction} & \noGeo & 61.3 & 54.7 & 59.4 & 54.2 & 85.6 & \best 47.5 & \best 107.3 & \best 31 & \best 60.9 \\
3DRS~\cite{huang3DRSMLLMsNeed2025} & \noGeo & 62.9 & 56.1 & 60.4 & 54.9 & 86.1 & 41.6 & 104.8 & 30.3 & 60.6 \\
\textbf{\ours{}} & \noGeo & \best 63.3 & \best 56.4 & \best 61.1 & \best 55.6 & \best 86.7 & 42.2 & 104.9 & 30.4 & 60.8 \\
\bottomrule
\end{tabular}
}
\vspace{-1em}
\end{table}

\textbf{Distillation and dual pathway layers.}
\label{sec:ablation_layer}
We ablate the choice of distillation layer in \Cref{table:distill_layer}. The results show that distillation from early or middle layers underperforms the SFT-only baseline, suggesting that the VLM requires sufficient computational depth to capture geometric relations. In contrast, later-layer distillation improves performance, with the 20th layer achieving the best overall score. 
Also, \Cref{table:layer_separation} shows the impact of parameter separation layer in our dual pathway design: applying separation to layers 13--20 yields the best result among the evaluated choices.


\textbf{Robustness on VLM backbone.}
We additionally evaluate \ours{} on two representative VLM backbones, InternVL3-8B~\cite{zhu2025internvl3} and Qwen2.5-VL-7B~\cite{bai2025qwen25vltechnicalreport}, using the same training setup as in \Cref{sec:vsi-experiments} except for the backbone. 
Under the controlled setting, as shown in \Cref{table:different_vlms}, \ours{} consistently outperforms feature distillation with further gains from Dual Pathway, consistent with the previous observation. 
It shows that the proposed relational distillation is robustly beneficial across diverse VLM backbones. 
More detailed results are presented in \Cref{table:vsi_3_MLLMs_detail}.



\textbf{3D scene understanding.}
To test whether MVRD generalizes beyond geometry-centric spatial QA, we train a separate model under the 3DRS scene-understanding protocol~\cite{huang3DRSMLLMsNeed2025}. 
This setting uses five standard 3D scene-understanding datasets: ScanRefer~\cite{chen2020scanrefer3dobjectlocalization}, Multi3DRefer~\cite{zhang2023multi3drefergroundingtextdescription}, Scan2Cap~\cite{chen2020scan2capcontextawaredensecaptioning}, ScanQA~\cite{azuma2022scanqa3dquestionanswering}, and SQA3D~\cite{ma2023sqa3dsituatedquestionanswering}, covering object grounding, dense captioning, and 3D question answering. 
Details are provided in \Cref{appendix:scene_understanding}.
\Cref{table:main_scene_understanding} shows that MVRD transfers well to 3D scene understanding tasks. 
It achieves the strongest results on object grounding, outperforming prior 3D scene understanding generalists on ScanRefer and Multi3DRefer, and improves over the feature-distillation baseline 3DRS~\cite{huang3DRSMLLMsNeed2025} across all metrics. 
These gains are consistent with our analysis in \Cref{sec:visual_rep_analysis}: MVRD injects multi-view geometric structure while better preserving text--visual alignment, which is especially important for grounding. 
\vspace{-0.5em}
\section{Conclusion}\label{sec:conclusion}
\vspace{-0.5em}
We addressed the problem of teaching vision-language models geometry without inflating inference cost or eroding pretrained vision-language alignment. We first identified a failure mode of feature distillation from geometry-grounded teachers: matching multi-view features pulls student representations away from the pretrained vision-language space, degrading linguistic grounding such as object-order tracking. We proposed \textbf{multi-view relational distillation (\ours{})}, which distills patch-wise cosine similarities across views, which can satisfy the geometric objective while staying near its pretrained representation. Across representative VLMs, \ours{} improves VSI-Bench accuracy over strong baselines, narrows the gap to a state-of-the-art feature fusion baseline, and transfers to 3D object grounding, dense captioning, and question answering. 

\textbf{Limitations.} Our study is conducted under moderate-scale training, and the method still induces some trade-offs. Future work could address this with large-scale training paired with efficient architectures that can encode both semantics and geometry.

\paragraph{Acknowledgments}
This work was in part supported by the Institute of Information \& Communications Technology Planning \& Evaluation (IITP) (RS-2022-II220926, RS-2026-25617720, and RS-2019-II190075) funded by the Korean government (MSIT).
We also thank Phillip Y. Lee and Duy M. H. Nguyen for their thoughtful discussions.

\bibliographystyle{unsrtnat}
\bibliography{main}

@String(CVPR= {IEEE Conf. Comput. Vis. Pattern Recog.})

@String(CVPR  = {CVPR})

@article{huang3DRSMLLMsNeed2025,
  title={3drs: Mllms need 3d-aware representation supervision for scene understanding},
  author={Huang, Xiaohu and Wu, Jingjing and Xie, Qunyi and Han, Kai},
  journal={Advances in Neural Information Processing Systems},
  volume={38},
  pages={67961--67988},
  year={2026}
}

@article{zhangSpatialStack2026,
  title={Spatialstack: Layered geometry-language fusion for 3d vlm spatial reasoning},
  author={Zhang, Jian and Zhou, Shijie and Liu, Bangya and Kadambi, Achuta and Fan, Zhiwen},
  journal={arXiv preprint arXiv:2603.27437},
  year={2026}
}

@inproceedings{zitkovich2023rt,
  title={Rt-2: Vision-language-action models transfer web knowledge to robotic control},
  author={Zitkovich, Brianna and Yu, Tianhe and Xu, Sichun and Xu, Peng and Xiao, Ted and Xia, Fei and Wu, Jialin and Wohlhart, Paul and Welker, Stefan and Wahid, Ayzaan and others},
  booktitle={Conference on Robot Learning},
  pages={2165--2183},
  year={2023},
  organization={PMLR}
}

@article{brohan2022rt,
  title={Rt-1: Robotics transformer for real-world control at scale},
  author={Brohan, Anthony and Brown, Noah and Carbajal, Justice and Chebotar, Yevgen and Dabis, Joseph and Finn, Chelsea and Gopalakrishnan, Keerthana and Hausman, Karol and Herzog, Alex and Hsu, Jasmine and others},
  journal={arXiv preprint arXiv:2212.06817},
  year={2022}
}

@article{driess2023palm,
  title={Palm-e: An embodied multimodal language model},
  author={Driess, Danny and Xia, Fei and Sajjadi, Mehdi SM and Lynch, Corey and Chowdhery, Aakanksha and Ichter, Brian and Wahid, Ayzaan and Tompson, Jonathan and Vuong, Quan and Yu, Tianhe and others},
  journal={arXiv preprint arXiv:2303.03378},
  year={2023}
}

@inproceedings{o2024open,
  title={Open x-embodiment: Robotic learning datasets and rt-x models: Open x-embodiment collaboration 0},
  author={O’Neill, Abby and Rehman, Abdul and Maddukuri, Abhiram and Gupta, Abhishek and Padalkar, Abhishek and Lee, Abraham and Pooley, Acorn and Gupta, Agrim and Mandlekar, Ajay and Jain, Ajinkya and others},
  booktitle={2024 IEEE International Conference on Robotics and Automation (ICRA)},
  pages={6892--6903},
  year={2024},
  organization={IEEE}
}

@book{craik1967nature,
  title={The nature of explanation},
  author={Craik, Kenneth James Williams},
  volume={445},
  year={1967},
  publisher={CUP Archive}
}

@book{marr1982vision,
  address = {New York, NY, USA},
  author = {Marr, David},
  description = {Vision},
  isbn = {0716715678},
  publisher = {Henry Holt and Co., Inc.},
  title = {Vision: A Computational Investigation into the Human Representation and Processing of Visual Information},
  year = 1982
}

@article{shepard1971mental,
  title={Mental rotation of three-dimensional objects},
  author={Shepard, Roger N and Metzler, Jacqueline},
  journal={Science},
  volume={171},
  number={3972},
  pages={701--703},
  year={1971},
  publisher={American Association for the Advancement of Science}
}

@book{johnson1983mental,
  title={Mental models: Towards a cognitive science of language, inference, and consciousness},
  author={Johnson-Laird, Philip Nicholas},
  number={6},
  year={1983},
  publisher={Harvard University Press}
}

@article{goodale1992separate,
  title={Separate visual pathways for perception and action},
  author={Goodale, Melvyn A and Milner, A David},
  journal={Trends in neurosciences},
  volume={15},
  number={1},
  pages={20--25},
  year={1992},
  publisher={Elsevier}
}

@book{o1978hippocampus,
  title={The hippocampus as a cognitive map},
  author={O'keefe, John and Nadel, Lynn},
  year={1978},
  publisher={Oxford university press}
}

@inproceedings{xie2025vlms,
  title={Are VLMs Ready for Autonomous Driving? An Empirical Study from the Reliability, Data and Metric Perspectives},
  author={Xie, Shaoyuan and Kong, Lingdong and Dong, Yuhao and Sima, Chonghao and Zhang, Wenwei and Chen, Qi Alfred and Liu, Ziwei and Pan, Liang},
  booktitle={Proceedings of the IEEE/CVF International Conference on Computer Vision},
  pages={6585--6597},
  year={2025}
}

@article{bai2025qwen25vltechnicalreport,
      title={Qwen2.5-VL Technical Report}, 
      author={Shuai Bai and Keqin Chen and Xuejing Liu and Jialin Wang and Wenbin Ge and Sibo Song and Kai Dang and Peng Wang and Shijie Wang and Jun Tang and Humen Zhong and Yuanzhi Zhu and Mingkun Yang and Zhaohai Li and Jianqiang Wan and Pengfei Wang and Wei Ding and Zheren Fu and Yiheng Xu and Jiabo Ye and Xi Zhang and Tianbao Xie and Zesen Cheng and Hang Zhang and Zhibo Yang and Haiyang Xu and Junyang Lin},
      journal={arXiv preprint arXiv:2502.13923},
      year={2025}
}

@article{zhu2025internvl3,
  title={Internvl3: Exploring advanced training and test-time recipes for open-source multimodal models},
  author={Zhu, Jinguo and Wang, Weiyun and Chen, Zhe and Liu, Zhaoyang and Ye, Shenglong and Gu, Lixin and Tian, Hao and Duan, Yuchen and Su, Weijie and Shao, Jie and others},
  journal={arXiv preprint arXiv:2504.10479},
  year={2025}
}

@article{
zhang2025llavavideo,
title={{LL}a{VA}-Video: Video Instruction Tuning With Synthetic Data},
author={Yuanhan Zhang and Jinming Wu and Wei Li and Bo Li and Zejun MA and Ziwei Liu and Chunyuan Li},
journal={Transactions on Machine Learning Research},
issn={2835-8856},
year={2025},
note={}
}

@article{kumar2025questioning,
  title={Questioning representational optimism in deep learning: The fractured entangled representation hypothesis},
  author={Kumar, Akarsh and Clune, Jeff and Lehman, Joel and Stanley, Kenneth O},
  journal={arXiv preprint arXiv:2505.11581},
  year={2025}
}

@inproceedings{li2025sti,
  title={Sti-bench: Are mllms ready for precise spatial-temporal world understanding?},
  author={Li, Yun and Zhang, Yiming and Lin, Tao and Liu, XiangRui and Cai, Wenxiao and Liu, Zheng and Zhao, Bo},
  booktitle={Proceedings of the IEEE/CVF International Conference on Computer Vision},
  pages={5622--5632},
  year={2025}
}

@inproceedings{zhou2025vlm4d,
  title={Vlm4d: Towards spatiotemporal awareness in vision language models},
  author={Zhou, Shijie and Vilesov, Alexander and He, Xuehai and Wan, Ziyu and Zhang, Shuwang and Nagachandra, Aditya and Chang, Di and Chen, Dongdong and Wang, Xin Eric and Kadambi, Achuta},
  booktitle={Proceedings of the IEEE/CVF international conference on computer vision},
  pages={8600--8612},
  year={2025}
}

@inproceedings{yang2025thinking,
  title={Thinking in space: How multimodal large language models see, remember, and recall spaces},
  author={Yang, Jihan and Yang, Shusheng and Gupta, Anjali W and Han, Rilyn and Fei-Fei, Li and Xie, Saining},
  booktitle={Proceedings of the Computer Vision and Pattern Recognition Conference},
  pages={10632--10643},
  year={2025}
}

@article{hu2025g,
  title={G2VLM: Geometry Grounded Vision Language Model with Unified 3D Reconstruction and Spatial Reasoning},
  author={Hu, Wenbo and Lin, Jingli and Long, Yilin and Ran, Yunlong and Jiang, Lihan and Wang, Yifan and Zhu, Chenming and Xu, Runsen and Wang, Tai and Pang, Jiangmiao},
  journal={arXiv preprint arXiv:2511.21688},
  year={2025}
}

@inproceedings{el2024probing,
  title={Probing the 3d awareness of visual foundation models},
  author={El Banani, Mohamed and Raj, Amit and Maninis, Kevis-Kokitsi and Kar, Abhishek and Li, Yuanzhen and Rubinstein, Michael and Sun, Deqing and Guibas, Leonidas and Johnson, Justin and Jampani, Varun},
  booktitle={Proceedings of the IEEE/CVF Conference on Computer Vision and Pattern Recognition},
  pages={21795--21806},
  year={2024}
}

@article{zheng2025learning,
  title={Learning from videos for 3d world: Enhancing mllms with 3d vision geometry priors},
  author={Zheng, Duo and Huang, Shijia and Li, Yanyang and Wang, Liwei},
  journal={arXiv preprint arXiv:2505.24625},
  year={2025}
}

@article{zhang2026spatialstack,
  title={SpatialStack: Layered Geometry-Language Fusion for 3D VLM Spatial Reasoning},
  author={Zhang, Jiang and Zhou, Shijie and Liu, Bangya and Kadambi, Achuta and Fan, Zhiwen},
  journal={arXiv preprint arXiv:2603.27437},
  year={2026}
}

@article{fan2025vlm,
  title={Vlm-3r: Vision-language models augmented with instruction-aligned 3d reconstruction},
  author={Fan, Zhiwen and Zhang, Jian and Li, Renjie and Zhang, Junge and Chen, Runjin and Hu, Hezhen and Wang, Kevin and Qu, Huaizhi and Zhou, Shijie and Wang, Dilin and others},
  journal={arXiv preprint arXiv:2505.20279},
  year={2025}
}

@inproceedings{wang2025vggt,
  title={Vggt: Visual geometry grounded transformer},
  author={Wang, Jianyuan and Chen, Minghao and Karaev, Nikita and Vedaldi, Andrea and Rupprecht, Christian and Novotny, David},
  booktitle={Proceedings of the Computer Vision and Pattern Recognition Conference},
  pages={5294--5306},
  year={2025}
}

@article{yang2025visual,
  title={Visual spatial tuning},
  author={Yang, Rui and Zhu, Ziyu and Li, Yanwei and Huang, Jingjia and Yan, Shen and Zhou, Siyuan and Liu, Zhe and Li, Xiangtai and Li, Shuangye and Wang, Wenqian and others},
  journal={arXiv preprint arXiv:2511.05491},
  year={2025}
}

@inproceedings{yang2025cambrian,
  title={Cambrian-s: Towards spatial supersensing in video},
  author={Yang, Shusheng and Yang, Jihan and Huang, Pinzhi and Brown II, Ellis L and Yang, Zihao and Yu, Yue and Tong, Shengbang and Zheng, Zihan and Xu, Yifan and Wang, Muhan and others},
  booktitle={The Fourteenth International Conference on Learning Representations},
  year={2025}
}

@article{cai2025scaling,
  title={Scaling spatial intelligence with multimodal foundation models},
  author={Cai, Zhongang and Wang, Ruisi and Gu, Chenyang and Pu, Fanyi and Xu, Junxiang and Wang, Yubo and Yin, Wanqi and Yang, Zhitao and Wei, Chen and Sun, Qingping and others},
  journal={arXiv preprint arXiv:2511.13719},
  year={2025}
}

@article{hinton2015distilling,
  title={Distilling the knowledge in a neural network},
  author={Hinton, Geoffrey and Vinyals, Oriol and Dean, Jeff},
  journal={arXiv preprint arXiv:1503.02531},
  year={2015}
}

@article{li2008unsupervised,
  title={Unsupervised natural experience rapidly alters invariant object representation in visual cortex},
  author={Li, Nuo and DiCarlo, James J},
  journal={science},
  volume={321},
  number={5895},
  pages={1502--1507},
  year={2008},
  publisher={American Association for the Advancement of Science}
}

@article{whittington2020tolman,
  title={The Tolman-Eichenbaum machine: unifying space and relational memory through generalization in the hippocampal formation},
  author={Whittington, James CR and Muller, Timothy H and Mark, Shirley and Chen, Guifen and Barry, Caswell and Burgess, Neil and Behrens, Timothy EJ},
  journal={Cell},
  volume={183},
  number={5},
  pages={1249--1263},
  year={2020},
  publisher={Elsevier}
}

@article{wiskott2002slow,
  title={Slow feature analysis: Unsupervised learning of invariances},
  author={Wiskott, Laurenz and Sejnowski, Terrence J},
  journal={Neural computation},
  volume={14},
  number={4},
  pages={715--770},
  year={2002},
  publisher={MIT Press}
}

@article{tommasi2012psychology,
  title={Psychology of spatial cognition},
  author={Tommasi, Luca and Laeng, Bruno},
  journal={Wiley Interdisciplinary Reviews: Cognitive Science},
  volume={3},
  number={6},
  pages={565--580},
  year={2012},
  publisher={Wiley Online Library}
}

@inproceedings{park2019relational,
  title={Relational knowledge distillation},
  author={Park, Wonpyo and Kim, Dongju and Lu, Yan and Cho, Minsu},
  booktitle={Proceedings of the IEEE/CVF conference on computer vision and pattern recognition},
  pages={3967--3976},
  year={2019}
}

@inproceedings{dai2017scannet,
  title={Scannet: Richly-annotated 3d reconstructions of indoor scenes},
  author={Dai, Angela and Chang, Angel X and Savva, Manolis and Halber, Maciej and Funkhouser, Thomas and Nie{\ss}ner, Matthias},
  booktitle={2017 IEEE conference on computer vision and pattern recognition (CVPR)},
  pages={2432--2443},
  year={2017},
  organization={IEEE}
}

@inproceedings{wang2025cut3r,
  title={Continuous 3d perception model with persistent state},
  author={Wang, Qianqian and Zhang, Yifei and Holynski, Aleksander and Efros, Alexei A and Kanazawa, Angjoo},
  booktitle={2025 IEEE/CVF Conference on Computer Vision and Pattern Recognition (CVPR)},
  pages={10510--10522},
  year={2025},
  organization={IEEE}
}

@article{hu2021loralowrankadaptationlarge,
  title={Lora: Low-rank adaptation of large language models},
  author={Hu, Edward J and Shen, Yelong and Wallis, Phillip and Allen-Zhu, Zeyuan and Li, Yuanzhi and Wang, Shean and Wang, Lu and Chen, Weizhu},
  journal={arXiv preprint arXiv:2106.09685},
  year={2021}
}

@article{li2024llavaonevisioneasyvisualtask,
  title={Llava-onevision: Easy visual task transfer},
  author={Li, Bo and Zhang, Yuanhan and Guo, Dong and Zhang, Renrui and Li, Feng and Zhang, Hao and Zhang, Kaichen and Zhang, Peiyuan and Li, Yanwei and Liu, Ziwei and others},
  journal={arXiv preprint arXiv:2408.03326},
  year={2024}
}

@article{huang2024chatscenebridging3dscene,
  title={Chat-Scene: Bridging 3D Scene and Large Language Models with Object Identifiers},
  author={Huang, Haifeng and Chen, Yilun and Wang, Zehan and Huang, Rongjie and Xu, Runsen and Wang, Tai and Liu, Luping and Cheng, Xize and Zhao, Yang and Pang, Jiangmiao and Zhao, Zhou},
  journal={arXiv preprint arXiv:2312.08168},
  year={2024}
}

@article{chen2024grounded3dllmreferenttokens,
  title={Grounded 3D-LLM with Referent Tokens},
  author={Chen, Yilun and Yang, Shuai and Huang, Haifeng and Wang, Tai and Xu, Runsen and Lyu, Ruiyuan and Lin, Dahua and Pang, Jiangmiao},
  journal={arXiv preprint arXiv:2405.10370},
  year={2024}
}

@article{zhu2024unifying3dvisionlanguageunderstanding,
  title={Unifying 3D Vision-Language Understanding via Promptable Queries},
  author={Zhu, Ziyu and Zhang, Zhuofan and Ma, Xiaojian and Niu, Xuesong and Chen, Yixin and Jia, Baoxiong and Deng, Zhidong and Huang, Siyuan and Li, Qing},
  journal={arXiv preprint arXiv:2405.11442},
  year={2024}
}

@article{zhu2025llava3dsimpleeffectivepathway,
  title={LLaVA-3D: A Simple yet Effective Pathway to Empowering LMMs with 3D-awareness},
  author={Zhu, Chenming and Wang, Tai and Zhang, Wenwei and Pang, Jiangmiao and Liu, Xihui},
  journal={arXiv preprint arXiv:2409.18125},
  year={2025}
}

@article{yu2025inst3dlmminstanceaware3dscene,
  title={Inst3D-LMM: Instance-Aware 3D Scene Understanding with Multi-modal Instruction Tuning},
  author={Yu, Hanxun and Li, Wentong and Wang, Song and Chen, Junbo and Zhu, Jianke},
  journal={arXiv preprint arXiv:2503.00513},
  year={2025}
}

@article{deng20253dllavageneralist3dlmms,
  title={3D-LLaVA: Towards Generalist 3D LMMs with Omni Superpoint Transformer},
  author={Deng, Jiajun and He, Tianyu and Jiang, Li and Wang, Tianyu and Dayoub, Feras and Reid, Ian},
  journal={arXiv preprint arXiv:2501.01163},
  year={2025}
}

@article{zheng2025video3dllmlearningpositionaware,
  title={Video-3D LLM: Learning Position-Aware Video Representation for 3D Scene Understanding},
  author={Zheng, Duo and Huang, Shijia and Wang, Liwei},
  journal={arXiv preprint arXiv:2412.00493},
  year={2025}
}

@article{openai2024gpt4ocard,
  title={GPT-4o System Card},
  author={OpenAI and Hurst, Aaron and Lerer, Adam and Goucher, Adam P. and Perelman, Adam and Ramesh, Aditya and Clark, Aidan and Ostrow, AJ and Welihinda, Akila and Hayes, Alan and Radford, Alec and others},
  journal={arXiv preprint arXiv:2410.21276},
  year={2024}
}

@article{comanici2025gemini25pushingfrontier,
  title={Gemini 2.5: Pushing the Frontier with Advanced Reasoning, Multimodality, Long Context, and Next Generation Agentic Capabilities},
  author={Comanici, Gheorghe and Bieber, Eric and Schaekermann, Mike and Pasupat, Ice and Sachdeva, Noveen and Dhillon, Inderjit and Blistein, Marcel and Ram, Ori and Zhang, Dan and Rosen, Evan and others},
  journal={arXiv preprint arXiv:2507.06261},
  year={2025}
}

@article{wang2025ross3dreconstructivevisualinstruction,
  title={Ross3D: Reconstructive Visual Instruction Tuning with 3D-Awareness},
  author={Wang, Haochen and Zhao, Yucheng and Wang, Tiancai and Fan, Haoqiang and Zhang, Xiangyu and Zhang, Zhaoxiang},
  journal={arXiv preprint arXiv:2504.01901},
  year={2025}
}

@article{li2025stibenchmllmsreadyprecise,
  title={STI-Bench: Are MLLMs Ready for Precise Spatial-Temporal World Understanding?},
  author={Li, Yun and Zhang, Yiming and Lin, Tao and Liu, Xiangrui and Cai, Wenxiao and Liu, Zheng and Zhao, Bo},
  journal={arXiv preprint arXiv:2503.23765},
  year={2025}
}

@article{zhou2025vlm4dspatiotemporalawarenessvision,
    title={VLM4D: Towards Spatiotemporal Awareness in Vision Language Models},
      author={Zhou, Shijie and Vilesov, Alexander and He, Xuehai and Wan, Ziyu and Zhang, Shuwang and Nagachandra, Aditya and Chang, Di and Chen, Dongdong and Wang, Xin Eric and Kadambi, Achuta},
      journal={arXiv preprint arXiv:2508.02095},
      year={2025}
}

@article{zhang2026flatlandspaceteachingvisionlanguage,
  title={From Flatland to Space: Teaching Vision-Language Models to Perceive and Reason in 3D},
  author={Zhang, Jiahui and Chen, Yurui and Zhou, Yanpeng and Xu, Yueming and Huang, Ze and Mei, Jilin and Chen, Junhui and Yuan, Yu-Jie and Cai, Xinyue and Huang, Guowei and Quan, Xingyue and Xu, Hang and Zhang, Li},
  journal={arXiv preprint arXiv:2503.22976},
  year={2026}
}

@article{yang2025visualspatialtuning,
  title={Visual Spatial Tuning},
  author={Yang, Rui and Zhu, Ziyu and Li, Yanwei and Huang, Jingjia and Yan, Shen and Zhou, Siyuan and Liu, Zhe and Li, Xiangtai and Li, Shuangye and Wang, Wenqian and Lin, Yi and Zhao, Hengshuang},
  journal={arXiv preprint arXiv:2511.05491},
  year={2025}
}

@article{yang2025cambriansspatialsupersensingvideo,
      title={Cambrian-S: Towards Spatial Supersensing in Video},
      author={Yang, Shusheng and Yang, Jihan and Huang, Pinzhi and Brown, Ellis and Yang, Zihao and Yu, Yue and Tong, Shengbang and Zheng, Zihan and Xu, Yifan and Wang, Muhan and Lu, Daohan and Fergus, Rob and LeCun, Yann and Fei-Fei, Li and Xie, Saining},
      journal={arXiv preprint arXiv:2511.04670},
      year={2025}
}

@article{cai2026scalingspatialintelligencemultimodal,
  title={Scaling Spatial Intelligence with Multimodal Foundation Models},
  author={Cai, Zhongang and Wang, Ruisi and Gu, Chenyang and Pu, Fanyi and Xu, Junxiang and Wang, Yubo and Yin, Wanqi and Yang, Zhitao and Wei, Chen and Sun, Qingping and Zhou, Tongxi and Li, Jiaqi and Pang, Hui En and Qian, Oscar and Wei, Yukun and Lin, Zhiqian and Shi, Xuanke and Deng, Kewang and Han, Xiaoyang and Chen, Zukai and Fan, Xiangyu and Deng, Hanming and Lu, Lewei and Pan, Liang and Li, Bo and Liu, Ziwei and Wang, Quan and Lin, Dahua and Yang, Lei},
  journal={arXiv preprint arXiv:2511.13719},
  year={2026}
}

@article{hu2025g2vlmgeometrygroundedvision,
      title={G$^2$VLM: Geometry Grounded Vision Language Model with Unified 3D Reconstruction and Spatial Reasoning}, 
      author={Wenbo Hu and Jingli Lin and Yilin Long and Yunlong Ran and Lihan Jiang and Yifan Wang and Chenming Zhu and Runsen Xu and Tai Wang and Jiangmiao Pang},
    journal={arXiv preprint arXiv:2511.21688},
    year={2025}
}

@article{chen2026think3dgeometricimagination,
      title={Think with 3D: Geometric Imagination Grounded Spatial Reasoning from Limited Views}, 
      author={Zhangquan Chen and Manyuan Zhang and Xinlei Yu and Xufang Luo and Mingze Sun and Zihao Pan and Xiang An and Yan Feng and Peng Pei and Xunliang Cai and Ruqi Huang},
      year={2026},
      journal={arXiv preprint arXiv:2510.18632},
}

@article{jeon2026visionalignedlatentreasoningmultimodal,
      title={Vision-aligned Latent Reasoning for Multi-modal Large Language Model}, 
      author={Byungwoo Jeon and Yoonwoo Jeong and Hyunseok Lee and Minsu Cho and Jinwoo Shin},
      year={2026},
      journal={arXiv preprint arXiv:2602.04476},
}

@article{dinov3,
      title={DINOv3}, 
      author={Oriane Siméoni and Huy V. Vo and Maximilian Seitzer and Federico Baldassarre and Maxime Oquab and Cijo Jose and Vasil Khalidov and Marc Szafraniec and Seungeun Yi and Michaël Ramamonjisoa and Francisco Massa and Daniel Haziza and Luca Wehrstedt and Jianyuan Wang and Timothée Darcet and Théo Moutakanni and Leonel Sentana and Claire Roberts and Andrea Vedaldi and Jamie Tolan and John Brandt and Camille Couprie and Julien Mairal and Hervé Jégou and Patrick Labatut and Piotr Bojanowski},
      year={2025},
      journal={arXiv preprint arXiv:2508.10104},
}

@article{bolya2025perceptionencoderbestvisual,
      title={Perception Encoder: The best visual embeddings are not at the output of the network}, 
      author={Daniel Bolya and Po-Yao Huang and Peize Sun and Jang Hyun Cho and Andrea Madotto and Chen Wei and Tengyu Ma and Jiale Zhi and Jathushan Rajasegaran and Hanoona Rasheed and Junke Wang and Marco Monteiro and Hu Xu and Shiyu Dong and Nikhila Ravi and Daniel Li and Piotr Dollár and Christoph Feichtenhofer},
      year={2025},
      journal={arXiv preprint arXiv:2504.13181},
}

@article{yeshwanth2023scannethighfidelitydataset3d,
      title={ScanNet++: A High-Fidelity Dataset of 3D Indoor Scenes}, 
      author={Chandan Yeshwanth and Yueh-Cheng Liu and Matthias Nießner and Angela Dai},
      year={2023},
      journal={arXiv preprint arXiv:2308.11417},
}

@article{baruch2022arkitscenesdiverserealworlddataset,
      title={ARKitScenes: A Diverse Real-World Dataset For 3D Indoor Scene Understanding Using Mobile RGB-D Data}, 
      author={Gilad Baruch and Zhuoyuan Chen and Afshin Dehghan and Tal Dimry and Yuri Feigin and Peter Fu and Thomas Gebauer and Brandon Joffe and Daniel Kurz and Arik Schwartz and Elad Shulman},
      year={2022},
      journal={arXiv preprint arXiv:2111.08897},
}

@article{chen2020scanrefer3dobjectlocalization,
      title={ScanRefer: 3D Object Localization in RGB-D Scans using Natural Language}, 
      author={Dave Zhenyu Chen and Angel X. Chang and Matthias Nießner},
      year={2020},
      journal={arXiv preprint arXiv:1912.08830},
}

@article{zhang2023multi3drefergroundingtextdescription,
      title={Multi3DRefer: Grounding Text Description to Multiple 3D Objects}, 
      author={Yiming Zhang and ZeMing Gong and Angel X. Chang},
      year={2023},
      journal={arXiv preprint arXiv:2309.05251},
}

@article{chen2020scan2capcontextawaredensecaptioning,
      title={Scan2Cap: Context-aware Dense Captioning in RGB-D Scans}, 
      author={Dave Zhenyu Chen and Ali Gholami and Matthias Nießner and Angel X. Chang},
      year={2020},
      journal={arXiv preprint arXiv:2012.02206},
}

@article{azuma2022scanqa3dquestionanswering,
      title={ScanQA: 3D Question Answering for Spatial Scene Understanding}, 
      author={Daichi Azuma and Taiki Miyanishi and Shuhei Kurita and Motoaki Kawanabe},
      year={2022},
      journal={arXiv preprint arXiv:2112.10482},
}

@article{ma2023sqa3dsituatedquestionanswering,
      title={SQA3D: Situated Question Answering in 3D Scenes}, 
      author={Xiaojian Ma and Silong Yong and Zilong Zheng and Qing Li and Yitao Liang and Song-Chun Zhu and Siyuan Huang},
      year={2023},
      journal={arXiv preprint arXiv:2210.07474},
}

@article{schult2023mask3dmasktransformer3d,
      title={Mask3D: Mask Transformer for 3D Semantic Instance Segmentation}, 
      author={Jonas Schult and Francis Engelmann and Alexander Hermans and Or Litany and Siyu Tang and Bastian Leibe},
      year={2023},
      journal={arXiv preprint arXiv:2210.03105},
}

@book{matthews2001short,
  title={A short history of structural linguistics},
  author={Matthews, Peter Hugoe},
  year={2001},
  publisher={Cambridge University Press}
}

@book{leinster2014basic,
  title={Basic category theory},
  author={Leinster, Tom},
  year={2014},
  publisher={Cambridge University Press}
}

@article{tong2024cambrian1fullyopenvisioncentric,
      title={Cambrian-1: A Fully Open, Vision-Centric Exploration of Multimodal LLMs}, 
      author={Shengbang Tong and Ellis Brown and Penghao Wu and Sanghyun Woo and Manoj Middepogu and Sai Charitha Akula and Jihan Yang and Shusheng Yang and Adithya Iyer and Xichen Pan and Ziteng Wang and Rob Fergus and Yann LeCun and Saining Xie},
      year={2024},
      journal={arXiv preprint arXiv:2406.16860}
}

@article{asadi2026mirageillusionvisualunderstanding,
      title={MIRAGE: The Illusion of Visual Understanding}, 
      author={Mohammad Asadi and Jack W. O'Sullivan and Fang Cao and Tahoura Nedaee and Kamyar Rajabalifardi and Fei-Fei Li and Ehsan Adeli and Euan Ashley},
      year={2026},
      journal={arXiv preprint arXiv:2603.21687},
}

@article{radford2021learningtransferablevisualmodels,
      title={Learning Transferable Visual Models From Natural Language Supervision}, 
      author={Alec Radford and Jong Wook Kim and Chris Hallacy and Aditya Ramesh and Gabriel Goh and Sandhini Agarwal and Girish Sastry and Amanda Askell and Pamela Mishkin and Jack Clark and Gretchen Krueger and Ilya Sutskever},
      year={2021},
      journal={arXiv preprint arXiv:2103.00020},
}

@article{zhai2023sigmoidlosslanguageimage,
      title={Sigmoid Loss for Language Image Pre-Training}, 
      author={Xiaohua Zhai and Basil Mustafa and Alexander Kolesnikov and Lucas Beyer},
      year={2023},
      journal={arXiv preprint arXiv:2303.15343},
}

@article{zhang2026videorepa,
  title={Videorepa: Learning physics for video generation through relational alignment with foundation models},
  author={Zhang, Xiangdong and Liao, Jiaqi and Zhang, Shaofeng and Meng, Fanqing and Wan, Xiangpeng and Yan, Junchi and Cheng, Yu},
  journal={Advances in Neural Information Processing Systems},
  volume={38},
  pages={122647--122676},
  year={2026}
}

@inproceedings{bhowmik2026moalign,
  title={Moalign: Motion-centric representation alignment for video diffusion models},
  author={Bhowmik, Aritra and Korzhenkov, Denis and Snoek, Cees G and Habibian, Amirhossein and Ghafoorian, Mohsen},
  booktitle={International Conference on Learning Representations},
  volume={2026},
  pages={122379--122397},
  year={2026}
}

@inproceedings{addepalli2024leveraging,
  title={Leveraging vision-language models for improving domain generalization in image classification},
  author={Addepalli, Sravanti and Asokan, Ashish Ramayee and Sharma, Lakshay and Babu, R Venkatesh},
  booktitle={2024 IEEE/CVF Conference on Computer Vision and Pattern Recognition (CVPR)},
  pages={23922--23932},
  year={2024},
  organization={IEEE}
}

@inproceedings{shah2020pitfalls,
  title={The Pitfalls of Simplicity Bias in Neural Networks},
  author={Shah, Harshay and Tamuly, Kaustav and Raghunathan, Aditi and Jain, Prateek and Netrapalli, Praneeth},
  booktitle={Advances in Neural Information Processing Systems},
  volume={33},
  year={2020},
}

@inproceedings{fu2024blink,
  title={Blink: Multimodal large language models can see but not perceive},
  author={Fu, Xingyu and Hu, Yushi and Li, Bangzheng and Feng, Yu and Wang, Haoyu and Lin, Xudong and Roth, Dan and Smith, Noah A and Ma, Wei-Chiu and Krishna, Ranjay},
  booktitle={European Conference on Computer Vision},
  pages={148--166},
  year={2024},
  organization={Springer}
}

@inproceedings{wang2025muirbench,
  title={Muirbench: A comprehensive benchmark for robust multi-image understanding},
  author={Wang, Fei and Fu, Xingyu and Huang, James Y and Li, Zekun and Liu, Qin and Liu, Xiaogeng and Ma, Mingyu Derek and Xu, Nan and Zhou, Wenxuan and Zhang, Kai and others},
  booktitle={International Conference on Learning Representations},
  volume={2025},
  pages={62624--62650},
  year={2025}
}


\newpage
\newpage
\appendix

\section{Proofs}
\label{appendix:proofs}

\begin{proof}[\textbf{Proof of \Cref{thm:fd_misalignment}}]
Since $\ell_{\mathsf{f}}$ is a distance metric, its minimum is zero and is attained if and only if the student feature equals the teacher feature, that is $\mathbf{v}_{\mathsf{f}}^\ast=\bar{\mathbf{g}}$, which proves the first part. For the second part, the Frobenius norm property $\|\mathbf{x}-\mathbf{y}\|^2=\|\mathbf{x}\|^2-2f(\mathbf{x},\mathbf{y})+\|\mathbf{y}\|^2$ gives
\begin{equation}
\|\bar{\mathbf{g}}-\mathbf{w}\|^2-\|\mathbf{v}-\mathbf{w}\|^2=\|\bar{\mathbf{g}}\|^2-\|\mathbf{v}\|^2+2\bigl(f(\mathbf{v},\mathbf{w})-f(\bar{\mathbf{g}},\mathbf{w})\bigr).
\end{equation}
Since each row of $\bar{\mathbf{g}}$ and $\mathbf{v}$ is $\ell_2$-normalized, $\|\bar{\mathbf{g}}\|=\|\mathbf{v}\|=\sqrt{N}$, so
\begin{equation}
f(\mathbf{v},\mathbf{w})-f(\bar{\mathbf{g}},\mathbf{w})=\tfrac{1}{2}\bigl(\|\bar{\mathbf{g}}-\mathbf{w}\|^2-\epsilon^2\bigr).
\end{equation}
By the triangle inequality, $\|\bar{\mathbf{g}}-\mathbf{w}\|\geq\delta-\epsilon$, hence
\begin{equation}
f(\mathbf{v},\mathbf{w})-f(\bar{\mathbf{g}},\mathbf{w})\geq\tfrac{1}{2}\bigl((\delta-\epsilon)^2-\epsilon^2\bigr)=\tfrac{1}{2}\delta(\delta-2\epsilon)>0,
\end{equation}
using $\delta>3\epsilon$ and $\mathbf{v}_{\mathsf{f}}^\ast=\bar{\mathbf{g}}$.
\end{proof}

\begin{lemma}[Orthogonal invariance of relational distillation]
\label{lem:rd_orthogonal_invariance}
For any $\mathbf{x}\in\mathbb{R}^{N\times d}$ and any orthogonal
matrix $\mathbf{Q}\in\mathbb{R}^{d\times d}$,
\begin{equation}
    \mathcal{L}_{\mathsf{r}}(\mathbf{x})
    = \mathcal{L}_{\mathsf{r}}(\mathbf{x}\mathbf{Q}).
\end{equation}
\end{lemma}

\begin{proof}
Orthogonality gives
\begin{equation}
    (\mathbf{x}\mathbf{Q})(\mathbf{x}\mathbf{Q})^\top
    = \mathbf{x}\mathbf{Q}\mathbf{Q}^\top\mathbf{x}^\top
    = \mathbf{x}\mathbf{x}^\top.
\end{equation}
Since $\mathcal{L}_{\mathsf{r}}$ depends on $\mathbf{x}$ only
through its Gram matrix, its value is unchanged.
\end{proof}

For any relational minimizer $\mathbf{v}_{\mathsf{r}}$,
\Cref{lem:rd_orthogonal_invariance} gives
$\mathcal{L}_{\mathsf{r}}(\mathbf{v}_{\mathsf{r}}\mathbf{Q})
=\mathcal{L}_{\mathsf{r}}(\mathbf{v}_{\mathsf{r}})=0$
for every orthogonal $\mathbf{Q}$. These transformations also
preserve row norms, yielding equally optimal feasible
representations with different global orientations.
\Cref{ass:rd_near_initialization} further requires the selected optimum to lie close to $\mathbf{v}$.
The alignment comparison below uses this additional assumption.

\begin{proof}[\textbf{Proof of \Cref{thm:rd_alignment}}]
By \Cref{ass:rd_near_initialization}, the distillation procedure
selects an optimum $\mathbf{v}_{\mathsf{r}}^\ast$ satisfying
$\rho=\|\mathbf{v}_{\mathsf{r}}^\ast-\mathbf{v}\|<\epsilon$.
Since the feature rows are $\ell_2$-normalized,
$\|\mathbf{v}_{\mathsf{r}}^\ast\|=\|\mathbf{v}\|=\sqrt{N}$.
The Frobenius norm identity therefore gives
\begin{equation}
    f(\mathbf{v},\mathbf{w})
    -f(\mathbf{v}_{\mathsf{r}}^\ast,\mathbf{w})
    =\tfrac{1}{2}\bigl(
      \|\mathbf{v}_{\mathsf{r}}^\ast-\mathbf{w}\|^2
      -\epsilon^2\bigr).
\end{equation}
By the triangle inequality,
$\|\mathbf{v}_{\mathsf{r}}^\ast-\mathbf{w}\|\leq\rho+\epsilon$,
so
\begin{equation}
    f(\mathbf{v},\mathbf{w})
    -f(\mathbf{v}_{\mathsf{r}}^\ast,\mathbf{w})
    \leq\tfrac{1}{2}\rho(\rho+2\epsilon).
\end{equation}
Combining this with \Cref{thm:fd_misalignment} yields
\begin{align}
    f(\mathbf{v}_{\mathsf{r}}^\ast,\mathbf{w})
    -f(\mathbf{v}_{\mathsf{f}}^\ast,\mathbf{w})
    &\geq\tfrac{1}{2}\delta(\delta-2\epsilon)
         -\tfrac{1}{2}\rho(\rho+2\epsilon) \\
    &=\tfrac{1}{2}(\delta+\rho)(\delta-\rho-2\epsilon).
\end{align}
Under \Cref{ass:relative_gap}, $\delta>3\epsilon$.
Together with $\rho<\epsilon$, this implies
$\delta-\rho-2\epsilon>\epsilon-\rho>0$, proving the claim.
\end{proof}

\section{Implementation Details}
\label{appendix:impl}

\subsection{Text-based Object Detection Precision}
\label{appendix:text-detection}

We use text-based object detection precision as a quantitative probe of
vision--language alignment in the VLM's internal visual representation.
We select visual patches with high cosine similarity to the object text token and measure the fraction that belong to the ground-truth patch set.
The metric is reported in \Cref{fig:compare_llava_3drs}(b) and \Cref{fig:compare_relation_feature_distill}(b) to assess the vision-language alignment of the distilled and base models.

\paragraph{Question selection.}
We restrict the probe to object nouns from the \emph{Object Size} task of VSI-Bench~\cite{yang2025thinking}, as each question refers to a single object class under a predefined format, making object noun extraction straightforward.
We further keep only those questions whose underlying scene video originates from
ScanNet~\cite{dai2017scannet}, so that ground-truth 3D bounding boxes,
camera poses, and depth maps are all available. To investigate the performance drop caused by distillation, we select questions on which SFT outperforms the distillation methods. This yields 42 questions from 42 scenes.

\paragraph{Forward pass and feature extraction.}
For each question, we uniformly sample 32 frames from the scene video and
feed them, together with the question text, into the VLM. At layer~$\ell$
we extract: (i) the visual-token hidden states
$\{H_i^{(\ell)} \in \mathbb{R}^{h_v \times w_v \times d_v}\}_{i=1}^{32}$,
and (ii) the hidden states of the subword tokens that spell out the
object's name in the prompt,
$\{e_j \in \mathbb{R}^{d_v}\}_{j=1}^{T}$.
We extract the object name from the fixed template used for VSI-Bench Object Size questions (e.g., the \texttt{<object>} slot in
\emph{``What is the length of the longest dimension \dots\ of the
\texttt{<object>}\dots\,?''}), and the corresponding subword tokens are
located by string matching against the tokenized prompt.
We average-pool these object-name token features into a single query
vector $\bar{e} = \tfrac{1}{T}\sum_{j=1}^{T} e_j$, which we refer to as
the \emph{noun token}.

\paragraph{Predicted patch set.}
We compute the cosine similarity
$s_{i,u,v} = \cos(\bar{e}, H_i^{(\ell)}[u,v])$
between the noun token and every visual patch across all
$32 \times h_v \times w_v$ positions in the scene, and let
$s_{\max} = \max_{i,u,v} s_{i,u,v}$.
Assuming that the queried object appears in the sampled frames, we define the predicted patch set adaptively as
\[
  \mathcal{P}_{\text{pred}}
  \;=\;
  \bigl\{(i,u,v) \,:\, s_{i,u,v} \,\geq\, 0.9\,s_{\max}\bigr\}.
\]
This relative threshold avoids the dependence on the scale of cosine similarity scores across different examples and model layers. 

\paragraph{Ground-truth patch set.}
For each frame $i$, we identify ground-truth object patches by comparing
the 3D coordinates extracted for each visual patch against the queried object's ground-truth 3D bounding box. 
We first unproject each valid depth pixel into world coordinates using
the depth map, camera intrinsics $K_i$, and camera pose/extrinsics provided
by ScanNet (\cref{appendix:backproj}). Pixels are then grouped according to
the image patch grid. For each patch, we compute the fraction of its valid
unprojected 3D points that fall inside the queried object's 3D bounding box,
using coordinate-wise bounds checking against the box's minimum and maximum
corners. A patch is labeled as a ground-truth patch if this fraction is at
least $50\%$. The ground-truth patch set $\mathcal{P}_{\mathrm{gt}}$ contains all ground-truth patches across frames.

\paragraph{Precision.}
For each question we compute
\[
  \text{Prec}
  \;=\;
  \frac{|\mathcal{P}_{\text{pred}} \cap \mathcal{P}_{\text{gt}}|}
       {|\mathcal{P}_{\text{pred}}|},
\]
and report the mean precision across all selected questions in \Cref{fig:compare_llava_3drs}, \Cref{fig:compare_relation_feature_distill}.

\subsection{3D Back-projection}
\label{appendix:backproj}

We use standard pinhole 3D back-projection on ScanNet using the camera
parameters and depth maps released with the dataset. For a pixel
$(u, v)$ with depth $d_{u,v}$ in a frame with intrinsics
$K \in \mathbb{R}^{3\times 3}$ and world-to-camera extrinsics
$[R \mid t] \in \mathbb{R}^{3\times 4}$, the 3D point in world
coordinates is
\[
  \mathbf{p}_{\text{cam}}
  = d_{u,v} \cdot K^{-1} \begin{bmatrix} u \\ v \\ 1 \end{bmatrix},
  \qquad
  \mathbf{p}_{\text{world}} = R^{\top}\!\bigl(\mathbf{p}_{\text{cam}} - t\bigr).
\]
Pixels with invalid (zero or missing) depth are discarded.

\subsection{Pearson Correlation}
\label{appendix:pearson}
We construct our relational distillation objective in Eq \eqref{eq:relational_distillation} using negative row-wise Pearson correlation between two relational representation matrices.
Given two real-valued vectors $a, b \in \mathbb{R}^{n}$, the Pearson
correlation coefficient is defined as:
\[
  \rho(a, b)
  \;=\;
  \frac{\sum_{k=1}^{n}(a_k - \bar{a})(b_k - \bar{b})}
       {\sqrt{\textstyle\sum_{k=1}^{n}(a_k - \bar{a})^2}\,
        \sqrt{\textstyle\sum_{k=1}^{n}(b_k - \bar{b})^2}},
\]
where $\bar{a}$ and $\bar{b}$ are the sample means. Pearson correlation
is invariant to positive affine rescaling
($\rho(\alpha a + \beta,\,b) = \rho(a,b)$ for $\alpha > 0$), which makes
it suitable for comparing two quantities measured on different scales, such as cosine distance and 3D Euclidean distance.

For two square matrices $A, B \in \mathbb{R}^{N \times N}$, we define the
\emph{row-wise} Pearson correlation as the average of per-row Pearson
correlations,
\[
  \rho_{\text{row}}(A, B) \;=\; \frac{1}{N}\sum_{i=1}^{N} \rho(A_{i,:},\,B_{i,:}).
\]

\subsection{Correlation between VGGT Multi-view Relations and 3D Distances}
\label{appendix:vggt-vs-3d}
\begin{figure}[h]
    \centering
    \includegraphics[width=0.8\textwidth]{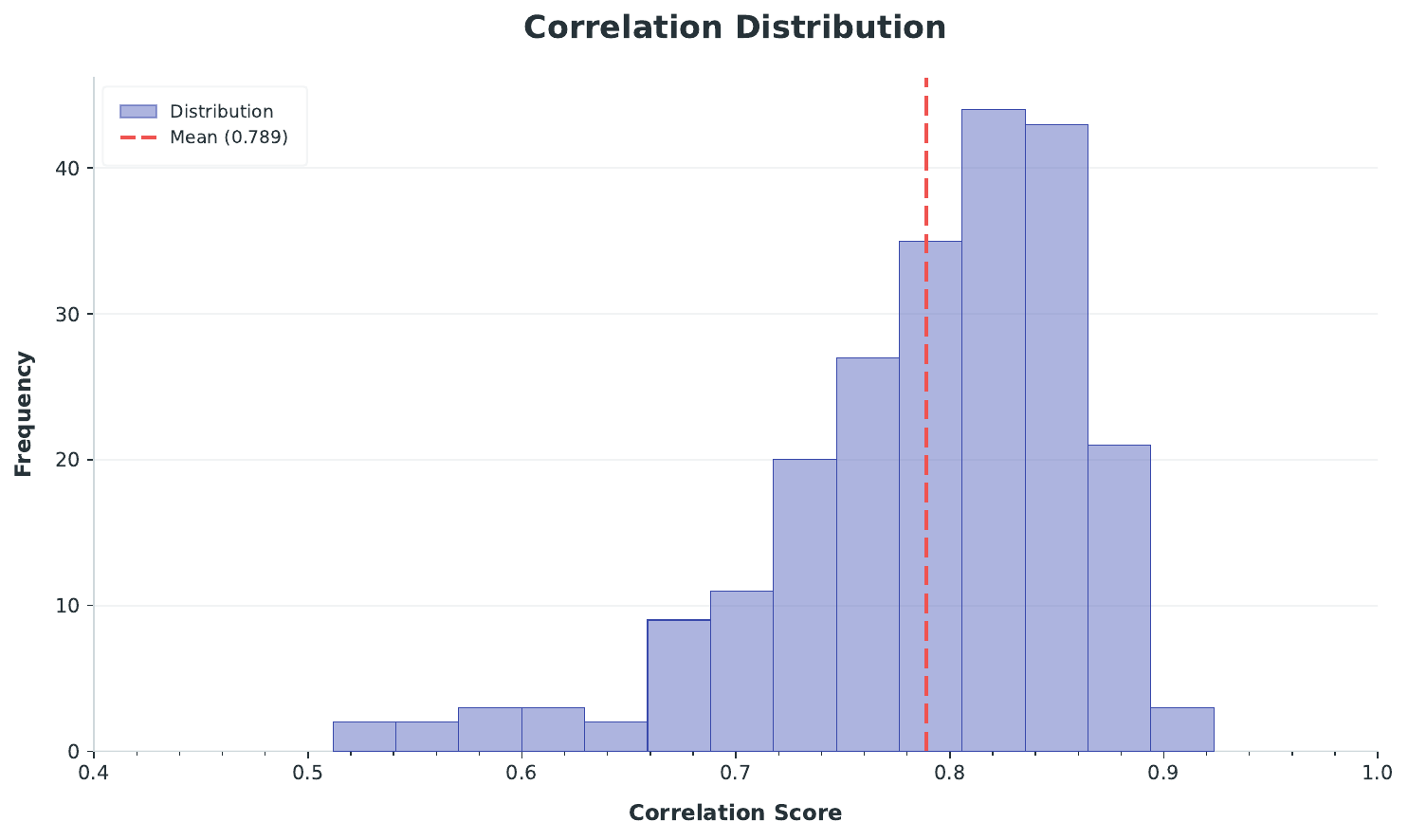}
    \caption{Row-wise Pearson correlation between VGGT cosine distances and 3D Euclidean distances on the ScanNet validation set.}
    \label{fig:correlation_vggt_3d_distance}
\end{figure}

We empirically verify that the multi-view similarity matrix $R^G$
extracted from VGGT~\cite{wang2025vggt} captures meaningful 3D structure by
comparing it to pairwise 3D Euclidean distances between the same
patches. For each scene video in the ScanNet validation set we proceed
as follows.
\begin{enumerate}
    \item \textbf{Frame sampling.} Uniformly sample 32 frames from the
        scene video and resize each to $378 \times 378$.
    \item \textbf{VGGT features.} Pass all 32 frames jointly through VGGT
        and extract patch-level feature maps of shape
        $32 \times 27 \times 27 \times d_g$ (patch size 14).
    \item \textbf{Multi-view similarity.} Flatten the feature maps to
        $H^G \in \mathbb{R}^{N \times d_g}$ with
        $N = 32 \cdot 27 \cdot 27 = 23{,}328$, $\ell_2$-normalize each
        row to obtain $\tilde{H}^G$, and compute the Gram matrix
        $R^G = \tilde{H}^G (\tilde{H}^G)^{\top} \in \mathbb{R}^{N \times N}$.
    \item \textbf{3D back-projection.} For each frame, back-project all
        pixels with valid depth into 3D world coordinates following
        \cref{appendix:backproj}, using the depth and camera
        parameters distributed with ScanNet.
    \item \textbf{Patch-level 3D centroids.} For each $14 \times 14$
        patch, average the 3D coordinates of its constituent
        valid-depth pixels to obtain a single 3D centroid per patch.
        We retain only patches that contain at least one valid-depth
        pixel; let $\mathcal{V} \subseteq \{1,\dots,N\}$ denote the
        index set of retained patches.
    \item \textbf{Pairwise 3D distances.} Stack the centroids of the
        retained patches into $P \in \mathbb{R}^{|\mathcal{V}| \times 3}$
        and compute the pairwise Euclidean distance matrix
        $D \in \mathbb{R}^{|\mathcal{V}| \times |\mathcal{V}|}$ with
        $D_{ij} = \lVert P_i - P_j \rVert_2$.
    \item \textbf{Correlation.} Restrict $R^G$ to the same row/column
        index set $\mathcal{V}$, treat $1 - R^G_{\mathcal{V},\mathcal{V}}$
        as a learned distance, and compute the row-wise Pearson
        correlation $\rho_{\text{row}}\bigl(1 - S^G_{\mathcal{V},\mathcal{V}},\,D\bigr)$
        as defined in \cref{appendix:pearson}.
\end{enumerate}

The mean row-wise Pearson correlation across ScanNet validation scenes is $\mathbf{0.79}$, with the detailed correlation distribution in \Cref{fig:correlation_vggt_3d_distance}.
This result supports using VGGT’s multi-view relations as a proxy for 3D proximity and as a distillation target when ground-truth 3D annotations are unavailable.


\section{Categorization of Spatial Reasoning Tasks}
\label{appendix:task_taxonomy}

To better diagnose the trade-off observed in our analysis (\Cref{3DRS_and_SFT,table:vsibench_baselines_comparison,table:ablation_representation_relation}), we partition the
eight VSI-Bench tasks into two categories according to the dominant type of
spatial reasoning each one demands:
\begin{enumerate}
    \item \textbf{Geometry-centered tasks}
          (\emph{Abs.\ Dist., Room Size, Rel.\ Dist., Rel.\ Dir., Route Plan}):
          tasks that require aggregating observations across frames into a coherent allocentric model of a scene and performing transformations between viewpoints to estimate the geometric relation in the scene.

    \item \textbf{Linguistic-grounding-centered tasks}
          (\emph{Obj.\ Count, Obj.\ Size, Appr.\ Order}):
          tasks whose answers are primarily determined by reliably grounding
          object categories in input frames, with the scene's geometric structure playing a secondary role.
\end{enumerate}

\paragraph{Tasks in geometry-centered group.}
\begin{itemize}
    \item \emph{Absolute Distance.} The model must combine views to infer the spatial arrangement of the two queried objects, then estimate their distance using commonsense knowledge of physical scale.

    \item \emph{Room Size.} Aggregate all input views into a global estimate, while there is no requirement for object grounding inside the scene.

    \item \emph{Relative Distance / Direction.} The model must combine cross-view observations and conduct viewpoint transformation to construct the partial scene that contains all mentioned objects. While linguistic cues specify which objects to focus on, these tasks require extensive viewpoint transformation to compare the distances/directions of different object regions to the reference object.

    \item \emph{Route Plan.} The model must track changes in position and viewing direction while reasoning about free space and obstacles.
\end{itemize}

\paragraph{Tasks in the linguistic-grounding-centered group.}
\begin{itemize}
    \item \emph{Object Count.} Requires detecting every instance of a
    named category and de-duplicating across frames based on visual cues rather than egocentric-allocentric viewpoint transformation.

    \item \emph{Object Size.} Requires precise object grounding and common-sense priors about object sizes, including the target object and other objects in the scene as references, while the geometric reasoning plays a secondary role.

    \item \emph{Appearance Order.} Focus on temporal grounding - identify when each object first appears and order the objects by those times - without any need for an
    allocentric scene model.
\end{itemize}

The two groups emphasize different capabilities: integrating scene geometry across views, and grounding object names in individual frames while tracking them over time.

\section{Training Setups}
\label{appendix:hyperparams}

\subsection{Spatial Reasoning Experimental Details}
\label{appendix:hyperparams-spatial}

\paragraph{Frame-sampling augmentation.}
Using the same sampled frames throughout training can encourage the student to memorize relations between fixed frame positions rather than learning the
underlying scene geometry, especially when the number of scene videos is limited (about 2000 videos in VLM-3R dataset~\citep{fan2025vlm}).
We mitigate this with a simple augmentation in the frame-sampling scheme. 
Given a scene video of $N$ frames, we partition the frames into 32 contiguous chunks of equal length $S = \lfloor N/32 \rfloor$.
At every training step, we draw an offset $r$ uniformly from the first
chunk,
$r \sim \mathrm{Uniform}\{0, 1, \dots, S-1\}$,
and use the 32 input frames at indices
\[
  \bigl\{\,r,\; r + S,\; r + 2S,\; \dots,\; r + 31S\,\bigr\}.
\]
This places exactly one frame in each chunk and preserves the uniform
stride along the temporal axis, while varying the visual information across training iterations. 
We found that a uniform interval between consecutive frames is important, as large temporal gaps introduce abrupt changes and reduce overlapping visual cues for scene understanding.
The augmentation is applied only during training; evaluation follows the deterministic uniform sampling.

\paragraph{Hyperparameters.}
\Cref{table:hyperparams-spatial} summarizes the training configuration
used for our spatial-reasoning experiments. 
Following VLM-3R~\citep{fan2025vlm}, we use a learning-rate schedule configured for five epochs and stop training after one epoch.
Training took 17 hours with 8 NVIDIA H200 GPUs.

\begin{table}[h]
\centering
\caption{Training hyperparameters for the spatial-reasoning experiments
on the VLM-3R~\cite{fan2025vlm} dataset.}
\label{table:hyperparams-spatial}
\begin{tabular}{ll}
\toprule
Hyperparameter & Value \\
\midrule
\multicolumn{2}{l}{\textit{Distillation}} \\
\quad Distillation loss weight $\gamma$ & $0.5$ \\
\midrule
\multicolumn{2}{l}{\textit{LoRA adapter}} \\
\quad LoRA enabled & True \\
\quad Rank $r$ & $128$ \\
\quad Alpha $\alpha$ & $256$ \\
\midrule
\multicolumn{2}{l}{\textit{Optimization}} \\
\quad Scheduled training epochs & 5 (stopped after epoch 1) \\
\quad Num GPUS & 8 \\
\quad Per-device batch size & 1 \\
\quad Gradient accumulation steps & 16 \\
\quad Peak learning rate & $2 \times 10^{-5}$ \\
\quad Weight decay & $0.0$ \\
\quad Warmup ratio & $0.03$ \\
\quad Learning-rate scheduler & cosine \\
\midrule
\multicolumn{2}{l}{\textit{System}} \\
\quad Numerical precision & BF16 (TF32 enabled) \\
\quad Maximum sequence length & $32{,}768$ tokens \\
\quad Gradient checkpointing & False \\
\bottomrule
\end{tabular}
\end{table}

\subsection{Scene Understanding Experimental Details}
\label{appendix:scene_understanding}

\begin{table}[t!]
\centering
\small
\caption{Datasets, evaluation splits, and metrics used for the 3D scene-understanding experiments.}
\label{tab:scene_understanding_protocol}
\begin{tabular}{l l l l}
\toprule
Dataset & Task type & Eval split & Metrics \\
\midrule
ScanRefer~\cite{chen2020scanrefer3dobjectlocalization}
& Single-object grounding
& Val.
& Acc@0.25, Acc@0.5 \\

Multi3DRefer~\cite{zhang2023multi3drefergroundingtextdescription}
& Multi-object grounding
& Val.
& F1@0.25, F1@0.5 \\

Scan2Cap~\cite{chen2020scan2capcontextawaredensecaptioning}
& Dense object captioning
& Val.
& C@0.5, B-4@0.5 \\

ScanQA~\cite{azuma2022scanqa3dquestionanswering}
& 3D question answering
& Val.
& CIDEr, EM \\

SQA3D~\cite{ma2023sqa3dsituatedquestionanswering}
& Situated 3D question answering
& Test
& EM \\
\bottomrule
\end{tabular}
\end{table}

\begin{table}[t!]
\centering
\small
\caption{Training configuration for the 3D scene-understanding experiments.}
\label{tab:scene_understanding_hyperparams}
\begin{tabular}{l l}
\toprule
Hyperparameter & Value \\
\midrule
Base model & LLaVA-Video-7B-Qwen2~\cite{zhang2025llavavideo} \\
Training protocol & 3DRS / Video-3D-LLM protocol~\cite{huang3DRSMLLMsNeed2025,zheng2025video3dllmlearningpositionaware} \\
Training datasets & ScanRefer, Multi3DRefer, Scan2Cap, ScanQA, SQA3D \\
Trainable setting & Full fine-tuning \\
Training epochs & 1 \\
Optimizer & Adam \\
Effective batch size & 16 \\
Language-model learning rate & $1\times10^{-5}$ \\
Vision-encoder learning rate & $2\times10^{-6}$ \\
Warmup ratio & 0.03 \\
Object proposals & Mask3D~\cite{schult2023mask3dmasktransformer3d} \\
Distillation layer & Last VLM layer \\
Hardware & 8 NVIDIA H200 GPUs \\
\bottomrule
\end{tabular}
\end{table}

For the scene-understanding results in Table~\ref{table:main_scene_understanding}, we train a separate model using the 3DRS protocol on a separate mixture of 3D scene-understanding datasets, rather than on the spatial-reasoning training data used for VSI-Bench. This experiment evaluates the applicability of \ours{} to object grounding, dense captioning, and 3D question answering.

\paragraph{Datasets.}
Following 3DRS~\cite{huang3DRSMLLMsNeed2025}, we train on the union of five standard 3D scene-understanding datasets, covering object grounding, dense captioning, and question answering. ScanRefer~\cite{chen2020scanrefer3dobjectlocalization} evaluates single-object grounding from natural-language descriptions. Multi3DRefer~\cite{zhang2023multi3drefergroundingtextdescription} extends this setting to descriptions that may refer to multiple objects. Scan2Cap~\cite{chen2020scan2capcontextawaredensecaptioning} evaluates dense captioning for an object specified by its 3D location within a scene. ScanQA~\cite{azuma2022scanqa3dquestionanswering} evaluates general 3D question answering, while SQA3D~\cite{ma2023sqa3dsituatedquestionanswering} evaluates situated question answering, where the model must reason from an embodied viewpoint in a 3D scene.

\vspace{-0.1em}
\paragraph{Evaluation protocol.}
We follow the evaluation protocol used by prior 3D scene-understanding generalists~\cite{zheng2025video3dllmlearningpositionaware,huang3DRSMLLMsNeed2025,wang2025ross3dreconstructivevisualinstruction}. We evaluate on the official test split of SQA3D and on the validation splits of ScanRefer, Multi3DRefer, Scan2Cap, and ScanQA. For ScanRefer, we report grounding accuracy at IoU thresholds of $0.25$ and $0.5$, denoted Acc@0.25 and Acc@0.5. For Multi3DRefer, we report F1 scores at the same IoU thresholds. For Scan2Cap, we report CIDEr and BLEU-4 under the IoU $0.5$ criterion, denoted C@0.5 and B-4@0.5. For ScanQA, we report CIDEr and Exact Match (EM). For SQA3D, we report EM as the metric.

\vspace{-0.1em}
\paragraph{Baselines.}
We compare with two families of 3D scene-understanding generalists. The first family uses dedicated 3D encoders to process point clouds or object-level 3D representations, including Chat-Scene~\cite{huang2024chatscenebridging3dscene}, Grounded 3D-LLM~\cite{chen2024grounded3dllmreferenttokens}, \textsc{PQ3D}~\cite{zhu2024unifying3dvisionlanguageunderstanding}, Inst3D-LMM~\cite{yu2025inst3dlmminstanceaware3dscene}, and 3D-LLaVA~\citep{deng20253dllavageneralist3dlmms}.
The second family builds on video VLMs augmented with 3D positional information or 3D-aware supervision, including LLaVA-3D~\cite{zhu2025llava3dsimpleeffectivepathway}, Video-3D-LLM~\cite{zheng2025video3dllmlearningpositionaware}, Ross3D~\cite{wang2025ross3dreconstructivevisualinstruction}, and 3DRS~\cite{huang3DRSMLLMsNeed2025}. When a baseline does not report a metric for a particular benchmark, we leave the corresponding entry blank in the main comparison table.

\vspace{-0.1em}
\paragraph{Object proposals and prediction format.}
For tasks requiring object-level grounding or captioning, we follow 3DRS~\cite{huang3DRSMLLMsNeed2025} and use object proposals extracted by Mask3D~\cite{schult2023mask3dmasktransformer3d}. Object-level visual features are obtained by aggregating the visual tokens associated with each 3D proposal. For grounding tasks, predictions are scored using the similarity between text features and proposal-level visual features, following the evaluation setting of prior work.

\vspace{-0.1em}
\paragraph{Training details.}
We use LLaVA-Video-7B-Qwen2~\cite{zhang2025llavavideo} as the base VLM and follow the full fine-tuning recipe of 3DRS~\cite{huang3DRSMLLMsNeed2025} and Video-3D-LLM~\cite{zheng2025video3dllmlearningpositionaware}. The model is trained for one epoch with Adam optimizer, an effective batch size of 16, and a warmup ratio of 0.03. We use a peak learning rate of $1\times10^{-5}$ for the language model and $2\times10^{-6}$ for the vision encoder. For \ours{}, multi-view relational distillation is applied at the last layer of the VLM. The model is trained on 8 NVIDIA H200 GPUs. The detailed configuration is shown in \Cref{tab:scene_understanding_hyperparams}.

\section{Additional Results}
\label{appendix:results}

\subsection{3DRS Baseline Performance on VSI-Bench}
\label{appendix:3drs_original}
\begin{table}[h]
\caption{Comparison of 3DRS~\citep{huang3DRSMLLMsNeed2025} and the SFT 
baselines~\citep{zheng2025learning,fan2025vlm} on VSI-Bench}
\label{table:3DRS_and_SFT_appendix}
\centering
\renewcommand{\arraystretch}{1.2}
\resizebox{\textwidth}{!}{%
\begin{tabular}{l | c | c c c c c c c c}
& &
\rotatebox{70}{Abs. Dist.} & 
\rotatebox{70}{Room Size} & 
\rotatebox{70}{Rel. Dist.} & 
\rotatebox{70}{Rel. Dir.} & 
\rotatebox{70}{Route Plan} & 
\rotatebox{70}{Obj. Count} & 
\rotatebox{70}{Obj. Size} & 
\rotatebox{70}{Appr. Order} \\
Methods & Avg. & 
\multicolumn{5}{c}{\cellcolor[HTML]{FFF2E5}\textbf{Geometric}} & 
\multicolumn{3}{c}{\cellcolor[HTML]{FFFFED}\textbf{Linguistic grounding}} \\
\hline
\rowcolor[HTML]{F2F7FB}
\textit{Training on VG LLM~\cite{zheng2025learning} S1 data} & & & & & & & & & \\
SFT~\cite{bai2025qwen25vltechnicalreport} & \textbf{49.8} & \textbf{36.0} & \textbf{59.7} & \textbf{45.8} & 38.9 & \textbf{30.4} & 68.6 & \textbf{57.9} & \textbf{60.2} \\
SFT + feature distillation (3DRS)~\cite{huang3DRSMLLMsNeed2025} & 45.9 & 34.8 & 56.6 & 40.9 & \textbf{43.2} & \textbf{30.4} & \textbf{68.7} & 53.6 & 39.2 \\
\hline
\rowcolor[HTML]{F2F7FB}
\textit{Training on VLM-3R~\cite{fan2025vlm} data} & & & & & & & & & \\
SFT~\cite{zhang2025llavavideo}& \textbf{57.7} & 43.6 & 63.7 & 64.9 & 68.9 & \textbf{40.7} & \textbf{70.6} & \textbf{70.8} & \textbf{38.5} \\
SFT + feature distillation (3DRS*)~\cite{huang3DRSMLLMsNeed2025} & 57.0 & \textbf{51.3} & \textbf{69.1} & \textbf{65.8} & \textbf{74.5} & 37.6 & 70.0 & 68.0 & 19.6 \\
\hline
\end{tabular}}
\end{table}
We report our reproduced results (Table~\ref{3DRS_and_SFT} in the main text) and the original results of 3DRS~\citep{huang3DRSMLLMsNeed2025} on VSI-Bench 
in Table~\ref{table:3DRS_and_SFT_appendix}. Compared to the SFT baseline using the 
same training dataset, 3DRS 
underperforms on most tasks and falls considerably short overall. 
The original and reproduced results both show lower overall accuracy than their corresponding SFT, confirming the detrimental effect of feature distillation.

\subsection{More Qualitative Results}
\label{appendix:qualitative}
\begin{figure}[h!]
    \centering
    \includegraphics[width=0.92\linewidth]{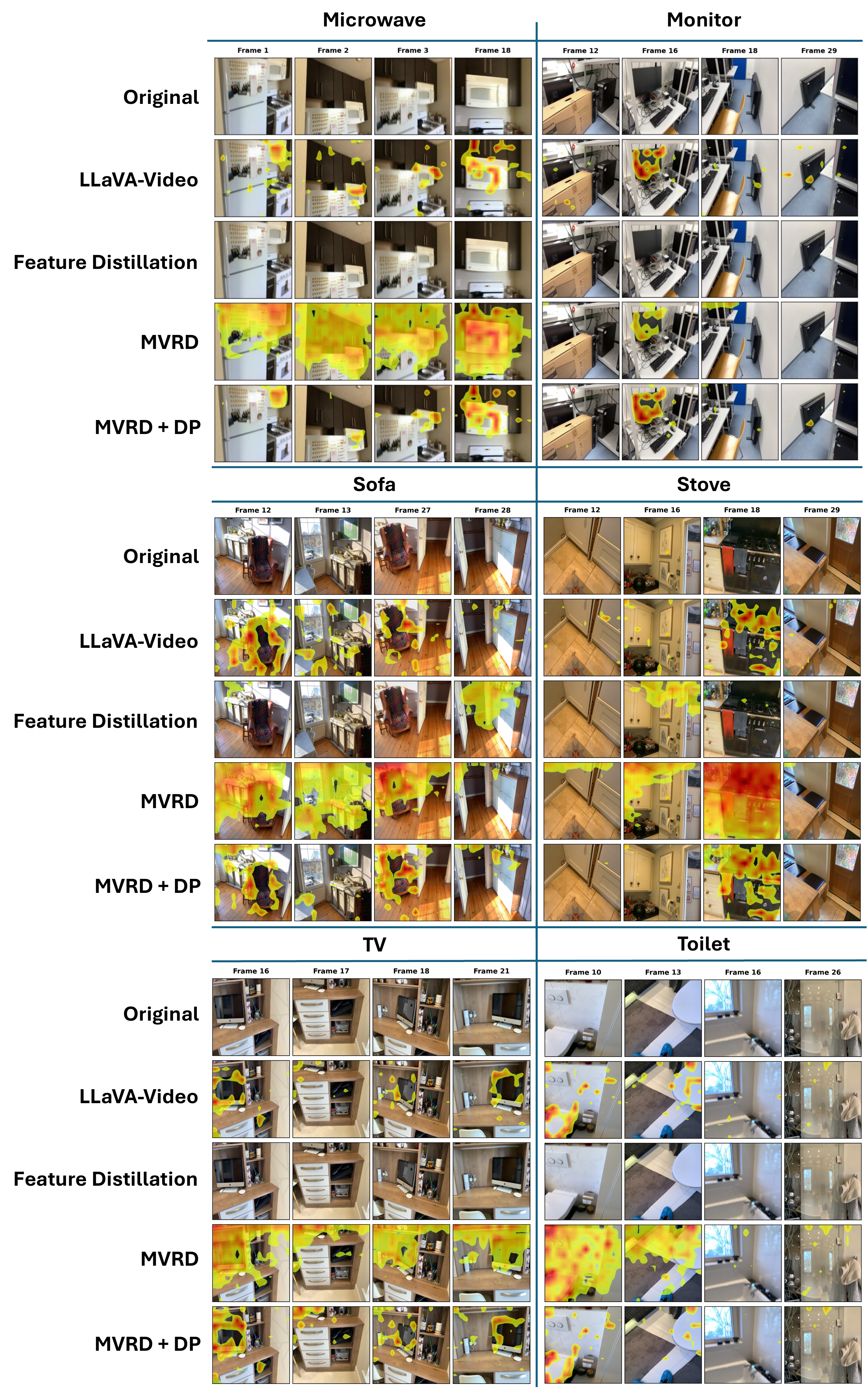}
    \caption{Additional text-to-patch similarity maps. \ours{} and \ours{} + Dual Pathway retain localized similarity patterns around the queried objects, whereas feature distillation diffuses or misaligns high-similarity regions.}
    \label{fig:more-text-vis}
    \vspace{-0.1in}
\end{figure}

We supplement \Cref{fig:compare_relation_feature_distill} of the main text with
additional qualitative comparisons (\Cref{fig:more-text-vis}) of text-to-patch similarity patterns
across all four model variants: the base LLaVA-Video, feature
distillation, our relational distillation (MVRD), and
MVRD + Dual Pathway under the same training setting using LoRA. 
For each example, we visualize the cosine similarity between the noun token of the queried object and the visual patches across all 32 input frames. 

\paragraph{Visualization detail.} For visualization, the similarity scores are min--max normalized separately for
each scene and each model variant as
\[
\tilde{s}_{m,n} = \frac{s_{m,n} - \min_n s_{m,n}}
{\max_n s_{m,n} - \min_n s_{m,n}},
\]
where $s_{m,n}$ denotes the cosine similarity for model $m$ and patch $n$
within a given scene. 
This normalization maps the scores of each model to
$[0,1]$, allowing the color scale to highlight the relative spatial
distribution of high-similarity patches within each model. 
Since the normalization is performed independently for each model and scene, the maps
should be interpreted as comparisons of localization patterns rather than
absolute similarity magnitudes across models. 
To improve visibility, we mask out the lowest 65\% of patch scores for each scene--model pair and visualize only the regions with relatively high text-to-patch similarity.

\paragraph{Observations.}
Compared with feature distillation, MVRD better preserves the concentration of high-similarity patches around the queried objects, which suggests that MVRD causes less disruption of the pre-trained feature space.
Furthermore, MVRD + Dual Pathway also retains the localized similarity patterns of the base model.

\subsection{More Quantitative Results}
\label{appendix:quantitative}

\begin{table}[h]
\centering
\caption{Per-task VSI-Bench results across VLM backbones. ``Feature.~Dist.''
denotes feature distillation; ``\ours{}'' is our relational
distillation. ``+\,DP'' marks Dual Pathway. Best
per column within each backbone block is in \textbf{bold}.}
\label{table:vsi_3_MLLMs_detail}
\small
\setlength{\tabcolsep}{3.5pt}
\renewcommand{\arraystretch}{1.2}
\resizebox{\linewidth}{!}{%
\begin{tabular}{l | l | c | c c c c c c c c}


& & &
\rotatebox{60}{Abs.\ Dist.}
& \rotatebox{60}{Room Size}
& \rotatebox{60}{Rel.\ Dist.}
& \rotatebox{60}{Rel.\ Dir.}
& \rotatebox{60}{Route Plan}
& \rotatebox{60}{Obj.\ Count}
& \rotatebox{60}{Obj.\ Size}
& \rotatebox{60}{Appr.\ Order} \\
Backbone & Method & Avg.
& \multicolumn{5}{c}{\cellcolor[HTML]{FFF2E5}\textbf{Geometric}}
& \multicolumn{3}{c}{\cellcolor[HTML]{FFFFED}\textbf{Linguistic Grounding}} \\
\midrule

\multirow{5}{*}{LLaVA-Video-7B}
 & SFT               & 57.7          & 43.6          & 63.7          & 64.9          & 68.9          & 40.7          & 70.6          & \textbf{70.8} & \textbf{38.5} \\
 & Feature.\ Dist.   & 58.5          & 48.3          & 68.4          & 61.4          & 75.7          & \textbf{44.9} & 71.4          & 67.8          & 30.1          \\
 & MVRD              & 59.9          & \textbf{50.5} & \textbf{69.3} & 62.7          & \textbf{80.0} & 40.7          & \textbf{72.9} & 68.3          & 35.1          \\
 & MVRD\,+\,DP       & \textbf{60.4} & 49.6          & 68.6          & 64.4          & 78.2          & 42.8          & 71.8          & 70.1          & 37.7          \\
\midrule
\multirow{4}{*}{InternVL3-8B}
 & SFT               & 50.3          & 41.8          & 52.9          & 57.5          & 53.2          & 25.8          & 70.8          & 67.5          & 33.3          \\
 & Feature \ Dist.   & 57.0          & 48.2          & 60.9          & 62.0          & 77.1          & 37.6          & 70.8          & 67.2          & 32.4          \\
 & MVRD              & 58.7          & 48.7          & \textbf{63.4} & 61.8          & \textbf{80.1} & \textbf{47.4} & \textbf{71.8} & 68.7          & 27.4          \\
 & MVRD\,+\, DP      & \textbf{59.9} & \textbf{50.3} & \textbf{63.4} & \textbf{63.5} & 76.8          & 44.3          & 71.7          & \textbf{68.9} & \textbf{40.1} \\
\midrule
\multirow{4}{*}{Qwen2.5-VL-7B}
 & SFT               & 56.1          & 47.0          & 53.7          & 61.8          & 73.8          & \textbf{45.9} & 70.5          & \textbf{70.7} & 25.1          \\
 & Feature \ Dist.   & 55.5          & 45.7          & 56.9          & 65.6          & 72.4          & 43.3          & 71.1          & 66.5          & 22.5          \\
 & MVRD              & 56.1          & 48.0          & 53.1          & \textbf{66.3} & 75.0          & 40.2          & \textbf{71.8} & 68.5          & 25.9          \\
 & MVRD\,+\,DP       & \textbf{59.8} & \textbf{50.9} & \textbf{62.2} & 63.2          & \textbf{76.2} & 44.3          & 71.7          & 68.9          & \textbf{40.1} \\
\bottomrule
\end{tabular}%
}
\end{table}

\paragraph{Per-task VSI-Bench results across VLM backbones.}
\Cref{table:vsi_3_MLLMs_detail} reports the full per-task breakdown of
\Cref{table:different_vlms} in the main text for InternVL3-8B and
Qwen2.5-VL-7B. We compare the Base VLM, supervised fine-tuning
(SFT), feature distillation, \ours{}, and \ours{} + Dual Pathway. Across both backbones, \ours{} consistently
outperforms \textit{feature distillation} on most of the tasks from both categories: \textit{geometric-centered} and\textit{ linguistic grounding-centered} QA, mirroring the
trend observed for LLaVA-Video-7B.

One notable exception is \textit{Appr. Order} on InternVL3, where \ours{} incurs a larger performance drop. We hypothesize that this backbone-specific behavior is related to InternVL3's video input format, which serializes frames as \texttt{"Frame-1: \textless image\textgreater, Frame-2: \textless image\textgreater,\ldots"}. Because appearance-order reasoning depends on associating visual content with these frame indices, enforcing cross-frame visual relations may partially interfere with this temporal association.

Importantly, our Dual Pathway variant, which separates the parameters used for relational learning, mitigates this degradation and provides more consistent improvements across tasks, with the largest gain on Qwen2.5-VL-7B (+\textbf{3.7\%} accuracy).


\paragraph{Ablation on multi-view relation learning.}
\label{appendix:relation-ablation}
We evaluate two design choices in Section~\ref{sec:method}: (i) the choice
of \emph{relation construction} - how the pairwise relation matrix is
computed in each feature space; and (ii) the choice of
\emph{matrix loss} - how the student and teacher relation matrices
are aligned. Concretely, given features
$\{f_i\}_{i=1}^{N}$ from either the VLM or the geometric teacher, we
consider two ways of forming the $N \times N$ relation matrix $R$:
\begin{itemize}
    \item \textbf{Cosine similarity} (used in MVRD):
        $R_{ij} = \cos(f_i, f_j)$,
        computed for both the $\ell_2$-normalized student and
        teacher features.
    \item \textbf{Euclidean distance}:
        $R_{ij} = \lVert f_i - f_j \rVert_2$,
        computed on the raw feature vectors in each space.
\end{itemize}
For aligning the student matrix $R^V$ with the teacher matrix $R^G$,
we consider:
\begin{itemize}
    \item \textbf{Row-wise Pearson correlation} (used in MVRD), computed between corresponding rows of the student and teacher relation matrices (\cref{appendix:pearson}).
    \item \textbf{Mean squared error (MSE)}, computed entry-wise as
        $\tfrac{1}{N^2}\sum_{i,j}(R^V_{ij} - R^G_{ij})^2$,
        which directly matches the absolute relation values.
\end{itemize}
The results are summarized in
\Cref{table:ablation-relation}. Other components - distillation layer,
LoRA configuration, and frame sampling - are kept the same as in the visual-spatial reasoning setup described in \cref{appendix:hyperparams-spatial}.

\begin{table}[t!]
\centering
\caption{Ablation on the construction of the cross-view relation matrix
and the distillation loss. VSI-Bench average accuracy ($\uparrow$).
The MVRD configuration is in \textbf{bold}.}
\label{table:ablation-relation}
\small
\begin{tabular}{lcc}
\toprule
 & Row-wise Pearson & MSE \\
\midrule
Pair-wise cosine similarity   & \textbf{59.9} & 58.4 \\
Pair-wise Euclidean distance  & 59.1          & 48.6 \\
\bottomrule
\end{tabular}
\end{table}

\noindent\textbf{Observations.} Row-wise Pearson outperforms MSE under
both relation constructions, with a much larger gap for Euclidean
distance ($+10.5$) than for cosine similarity ($+1.5$).
Euclidean distances depend on feature magnitudes, which may differ substantially between the student and teacher. MSE penalizes these scale differences even when their relational patterns agree, 
whereas row-wise Pearson correlation (used in \ours{}) is invariant to row-wise shifts and positive rescaling, allowing differences in scale and offset between student and teacher relation values.
This may explain the larger gain of row-wise Pearson correlation over MSE for Euclidean distance, since cosine similarity already normalizes feature magnitudes.

Likewise, cosine similarity outperforms Euclidean distance under both losses, suggesting that removing feature-magnitude differences makes relations easier to distill between the VLM and teacher.
The worst-performing combination (Euclidean + MSE, $48.6$) falls well below
the SFT baseline of $57.7$. Hence, this ablation indicates that both cosine similarity and Pearson correlation improve accuracy.


\end{document}